\documentclass[fleqn,10pt]{SelfArx} 

\usepackage[english]{babel} 

\usepackage[utf8]{inputenc} 
\usepackage[T1]{fontenc}    
\usepackage{hyperref}       
\usepackage{url}            
\usepackage{booktabs}       
\usepackage{amsfonts}       
\usepackage{nicefrac}       
\usepackage{microtype}      
\usepackage{fancyhdr}       
\usepackage{graphicx}       
\usepackage{multirow}
\usepackage{color}
\usepackage{soul}
\usepackage{caption}
\usepackage{subcaption}
\usepackage{amsmath}

\definecolor{color1}{RGB}{0,0,90} 
\definecolor{color2}{RGB}{0,20,20} 

\newcommand\bl[1]{\textcolor{blue}{#1}}

\usepackage{hyperref} 

\hypersetup{
	hidelinks,
	colorlinks,
	breaklinks=true,
	urlcolor=color2,
	citecolor=color1,
	linkcolor=color1,
	bookmarksopen=false,
	pdftitle={Title},
	pdfauthor={Author},
}

\JournalInfo{arXiv preprint} 
\Archive{in submission} 

\PaperTitle{Rescuing referral failures during automated diagnosis of domain-shifted medical images} 

\Authors{
Anuj Srivastava\textsuperscript{1},
Karm Patel\textsuperscript{1},
Pradeep Shenoy\textsuperscript{2},
Devarajan Sridharan\textsuperscript{1}*
}
\affiliation{\textsuperscript{1}\textit{Center for Neuroscience \& Computer Science and Automation, Indian Institute of Science, Bengaluru}} 
\affiliation{\textsuperscript{2}\textit{Google Research India, Bengaluru}} 
\affiliation{*\textbf{Corresponding author}: sridhar@iisc.ac.in} 

\newcommand{\keywordname}{Keywords} 

\newcommand{\abstracttext}{
The success of deep learning models deployed in the real world depends critically on their ability to generalize well across diverse data domains. Here, we address a fundamental challenge with selective classification during automated diagnosis with domain-shifted medical images. In this scenario, models must learn to avoid making predictions when label confidence is low, especially when tested with samples far removed from the training set (covariate shift). Such uncertain cases are typically referred to the clinician for further analysis and evaluation. Yet, we show that even state-of-the-art domain generalization approaches fail severely during referral when tested on medical images acquired from a different demographic or using a different technology. We examine two benchmark diagnostic medical imaging datasets exhibiting strong covariate shifts: i) diabetic retinopathy prediction with retinal fundus images and ii) multilabel disease prediction with chest X-ray images. We show that predictive uncertainty estimates do not generalize well under covariate shifts leading to non-monotonic referral curves, and severe drops in performance (up to $\sim$50\%) at high referral rates ($>$70\%). We evaluate novel combinations of robust generalization and {\it post hoc} referral approaches, that rescue these failures and achieve significant performance improvements, typically $>$10\%, over baseline methods. Our study identifies a critical challenge with referral in domain-shifted medical images and finds key applications in reliable, automated disease diagnosis.
}
\Abstract{\abstracttext}

\begin{document}

\emergencystretch 3em

\maketitle 

\tableofcontents 

\thispagestyle{empty} 


\section{Introduction}
\label{sec:intro}

\begin{figure*}[t]
\centering
\includegraphics[width=0.93\textwidth]{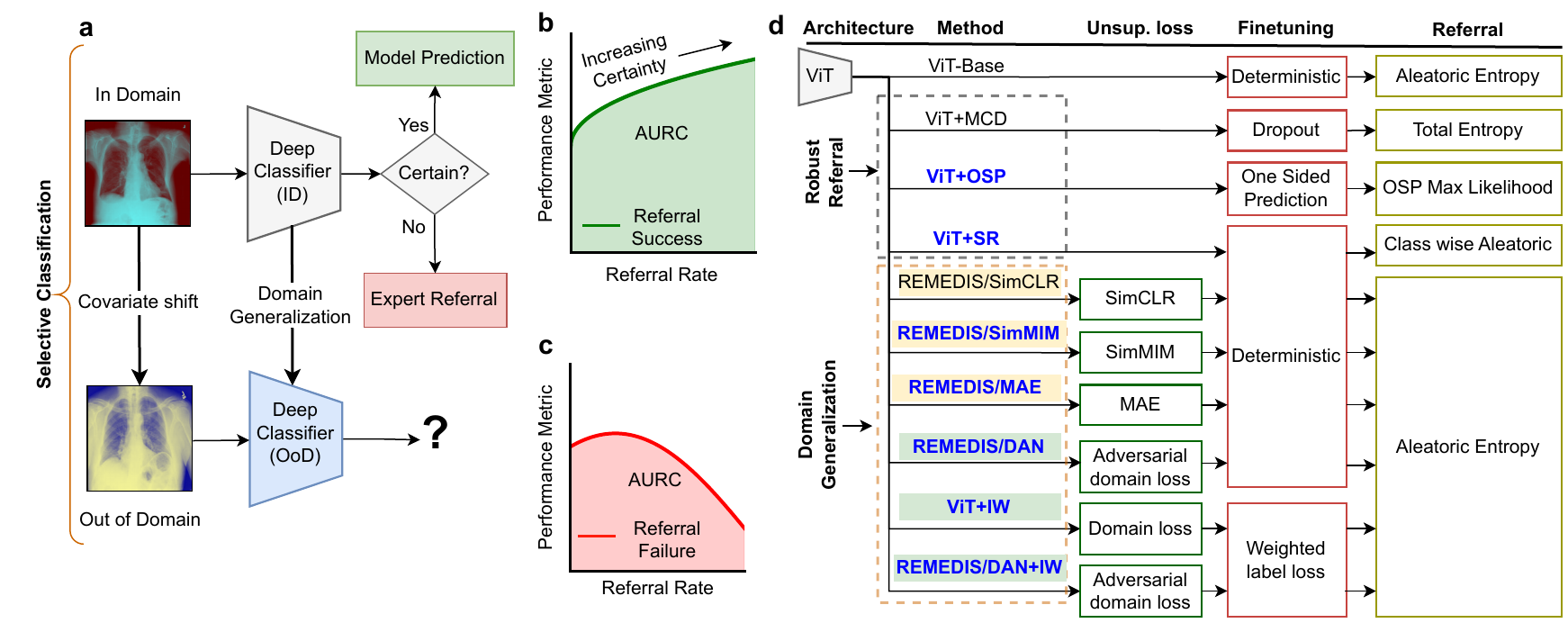} 
    \caption{
        { (a)} The challenge with selective classification under domain (covariate) shift. Although models may be well calibrated with in-domain (ID) data, enabling successful referral, they may be miscalibrated with out-of-domain (OOD) data resulting in referral failure.
        { (b)} and { (c)} Referral curves for success (b) and failure (c) cases. 
        Referral curves show a performance metric (e.g., AUROC or accuracy) as a function of referral rate.
        Successful referral curves exhibit monotonically increasing performance as the more uncertain samples are referred and the more confident ones, retained.
        Shaded region: Area Under the Referral Curve (AURC) used to quantify referral performance.
        {(d)} Robust referral under domain shift. Blue: Novel combinations proposed in this study. Yellow background shading: techniques with a self-supervised training objective; Green background shading: concurrent domain generalization and label prediction.
    }
    \label{fig:schematic}
\end{figure*}

Deep learning models deployed in real-world situations must be able to generalize readily when they encounter novel inputs. Yet, deep discriminative models do not perform well, typically, when tested on data that is far removed from the training distribution (domain shift). Previous domain generalization studies have explored at least one of two complementary types of domain shift: covariate shift and semantic shift. In covariate shift, the distribution of the input data changes between training and testing phases, whereas the target labels remain the same (e.g. a classifier trained on photographs deployed to categorize paintings, \cite{li2017deeper}). By contrast, in semantic shift, the distribution of input data and target labels assume additional degrees of freedom or meaning (e.g. a classifier trained to distinguish cars from boats deployed to classify trucks from ships). 

Here, we address a particular challenge with domain generalization for automated disease diagnosis with domain-shifted medical images. Specifically, we focus on challenges with selective classification under covariate shift. In this setting, a model must avoid making predictions when it is not confident about label assignment, especially when the input (test) data deviates significantly from its training distribution. Such ``uncertain'' instances are typically referred to a clinical expert for further diagnosis and decision-making. In an ideal scenario, this approach reduces the clinician's burden while allowing high-accuracy predictions only for the most confident samples. Yet, we show that model behavior is far from this ideal, even for state-of-the-art domain generalization approaches. 

Our study makes the following key contributions:
\begin{itemize}

\item We show that various classes of deep learning models, including i) those pretrained on large-scale, natural image datasets~\cite{dosovitskiy2021an, bit_kolesnikov2020big}, ii) state-of-the-art models designed for robust, data-efficient generalization ~\cite{azizi2023}, and even iii) those designed for accurate quantification of sample predictive uncertainty~\cite{band2021benchmarking}  fail -- often critically -- during selective classification with domain shifted images.

\item We illustrate these failures in large-scale datasets from two major medical imaging domains: i) diabetic retinopathy detection using retinal fundus images, and ii) multilabel prediction with chest X-ray images (Fig.~\ref{fig:schematic}a-c). 

\item Within these domains, we study datasets encompassing multiple kinds of covariate shifts, including technology shift (image acquisition with different make and model of medical equipment) and population shift (distinct demographics of patients).

\item We analyze reasons for these failures, and evaluate novel combinations of generalization and referral solutions that enable reliable selective classification under domain shift.

\end{itemize}

The results are relevant for reliable automated diagnostic medical imaging, especially when images are sourced from diverse populations or grades of imaging technologies.

\section{Challenges with referral under domain-shift}
\label{sec:motivn}
Robust referral with domain-shifted images requires a combination of efficient out-of-domain (OOD) generalization approaches and methods for reliable referral. The latter, in turn, demands accurate estimates of the model's predictive uncertainty. We demonstrate that even a combination of state-of-the-art generalization and uncertainty estimation approaches fails to achieve reliable referral on OOD data.

\subsection{Referral failures with domain shifted images}
A major challenge for any deep learning model developed for diagnostic medical imaging is the ability to generalize to out-of-domain data~\cite{domain_zech2018variable, domain_zhang2021empirical}. For example, models trained to accurately detect diabetic retinopathy (DR) with retinal fundus images from one population (in domain or ID) may fail to generalize well when tested with out-of-distribution (OOD) scans from a different population or demographic~\cite{band2021benchmarking}. Also, such models may fail to produce reliable estimates of predictive uncertainty to enable accurate referral with OOD data.

We illustrate this challenge with the Retina Benchmark \cite{band2021benchmarking}, which explores such referral failures with two kinds of domain shift: i) semantic shift and ii) covariate shift. In the semantic shift case, also termed ``Severity shift'', a model trained on retinal fundus images with (at most) moderate DR is evaluated on OOD images with severe or proliferative DR. In the covariate shift case, also termed ``Country shift'', a model trained on a DR dataset collected in the United States \cite{eyepacs} is evaluated on an OOD dataset collected in India \cite{aptos}. The latter case is particularly challenging because it includes a variety of attributes that differ between dataset domains, including demographics, ethnicity, imaging technology, image quality, comorbidities in each population, and differences in label distribution. We, therefore, 
demonstrate referral failures in this latter (Country shift) case. The underlying task is binary classification: whether a given test sample exhibits symptoms of diabetic retinopathy or not. For selective classification, the uncertainty in predicting a test sample's label is evaluated using the model's predictive entropy: 
\begin{equation}\label{eq:entropy}
    \mathcal{H} \left ( p_\theta \left ( y | x \right) \right ) = - \sum_{i=1}^C p_\theta(y=i|x) \log p_\theta(y=i|x)
\end{equation}
where $x$ is a test sample, $y$ denotes the model output, $C$ is the number of  classes, and $\theta$ are the model parameters.  

For referral, test samples with the highest entropies (highest predictive uncertainties) are referred first, while samples with the lowest entropies (lowest predictive uncertainties) are retained for classification. In the ideal scenario, model performance metrics, such as accuracy or area under the ROC (AUROC), should increase with increasing referral rates -- the proportion of cases referred to the expert.

\begin{figure}[ht]
    \centering

    \includegraphics[width=0.47\textwidth]{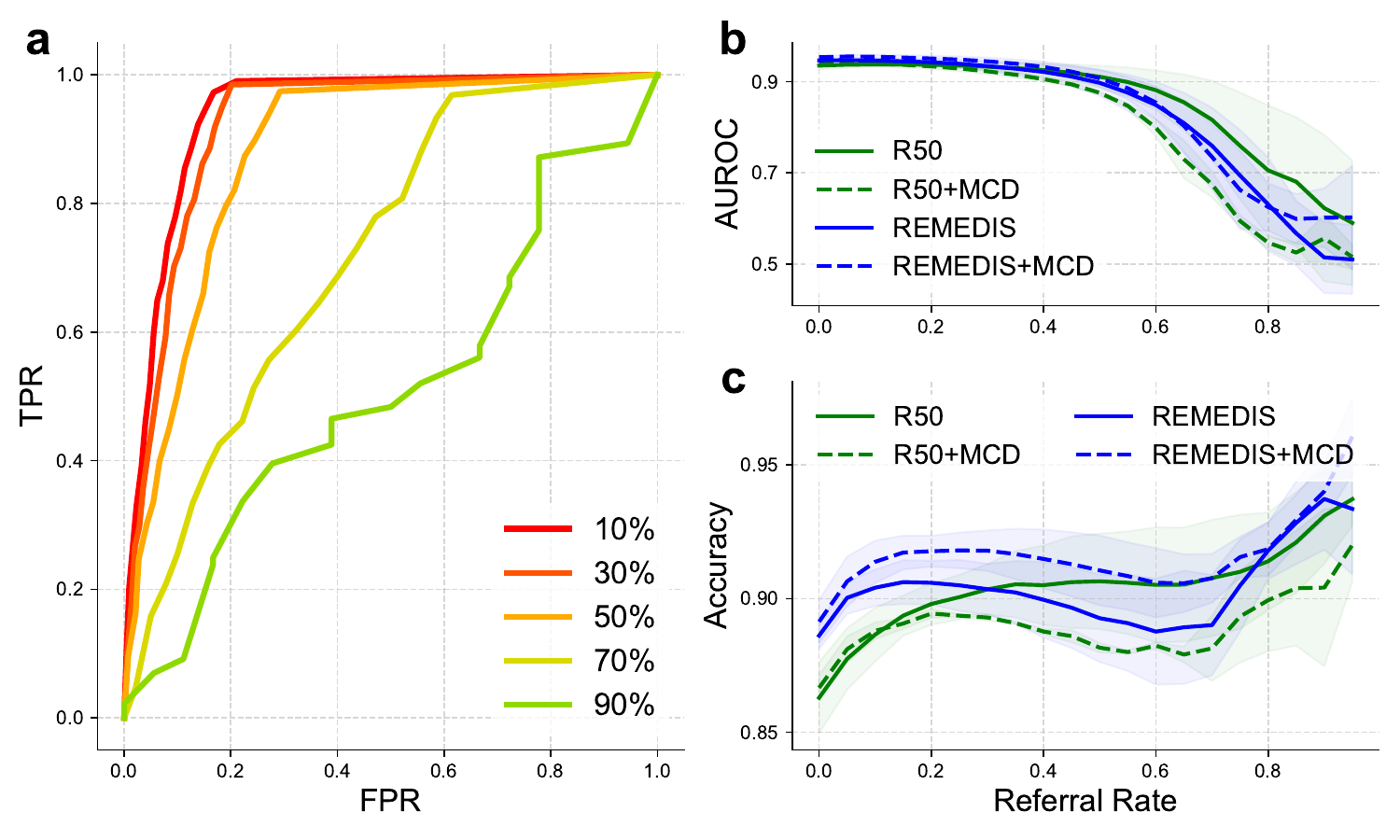}
    \caption{
        (a) ROCs for different referral rates (10-90\%) on the Diabetic Retinopathy OOD (APTOS) dataset with the baseline model (ResNet50).
        (b) and (c): Referral curves for AUROC (b) and Accuracy (c). See Fig. 1b for details.
    }
    \label{fig:rocs-refs-sec2}
\end{figure}
Yet, we observe a surprising reversal of this expected trend. We plot ROC curves for the baseline DR model (ResNet-50 architecture, {Appendix A.1}) for two different (high) referral rates of the OOD (APTOS) data (50\%, Fig. \ref{fig:rocs-refs-sec2}a, orange; and 70\%, Fig. \ref{fig:rocs-refs-sec2}a, yellow). AUROCs are lower for the 70th percentile sample with the most confident predictions ($\langle \mathcal{H} \rangle < 0.24$ nats, AUROC = 0.82), as compared to the 50th percentile sample with the less confident predictions, on average ($\langle \mathcal{H} \rangle < 0.37$ nats, AUROC = 0.91).

To further quantify this surprising behavior, we plot the  ``referral curve'' in which a performance metric (e.g. AUROC or accuracy) is plotted as a function of the proportion of OOD samples referred (Fig. \ref{fig:rocs-refs-sec2}b-c). The proportion of samples retained for evaluation by the model decreases along the x-axis: these represent samples with progressively higher model confidence. Therefore, AUROC or accuracy should increase monotonically with the referral rate. 
Yet, this is not the case. AUROC is high ($>$0.9) at zero referral rate and decreases progressively at higher referral rates (Fig. \ref{fig:rocs-refs-sec2}b green), with a sharp (precipitous) drop around a referral rate of around 60\%. Moreover, we observe a similarly non-monotonic behavior, albeit less pronounced, with the accuracy-referral rate curve also (Fig. \ref{fig:rocs-refs-sec2}c green).

We quantified these undesirable drops in performance with the area under the referral curve (AURC). We note that this metric is inversely and monotonically related to the area under the risk-coverage curve, among the most common measures of model reliability proposed in recent studies~\cite{aurc_ding2020revisiting}. For the OOD/APTOS dataset, AURC-AUROC was 0.86, considerably lower than the value at zero referral rate (0.94). Similarly, AURC-Accuracy was 0.90, comparable to its value at zero referral rates (0.86). 
By contrast, such failures are largely absent for in-domain (ID/EyePACS) data ({Appendix E.1}). For the ID/EyePACS dataset, AURC-AUROC was 0.91 and AURC-Accuracy was 0.95, considerably higher than their values at zero referral rates (0.87 and 0.89, respectively).

\subsection{OOD generalization with REMEDIS}
\begin{figure}[ht]
    \centering

    \includegraphics[width=0.47\textwidth]{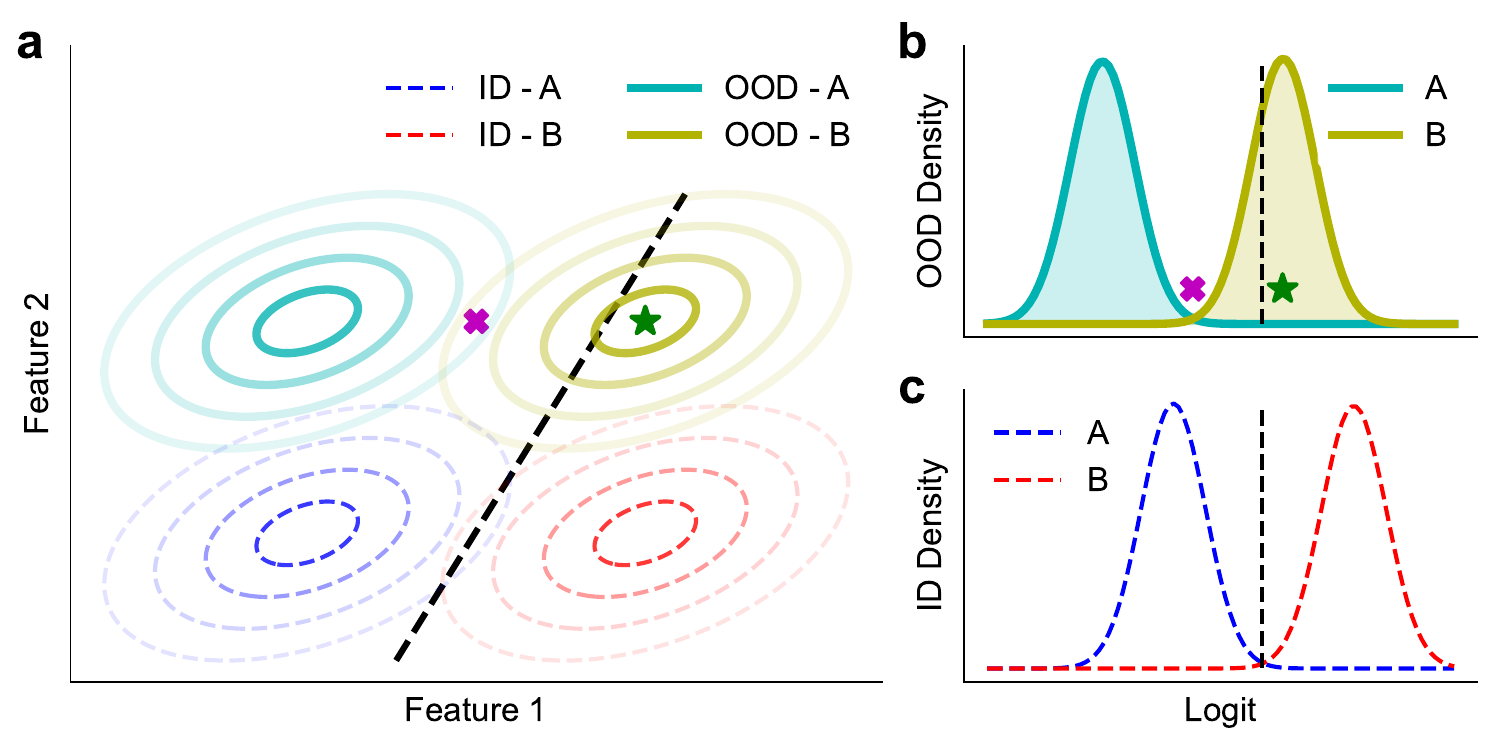}
    
    \caption{
    Simulated example. 
    (a) A classifier fitted on ID data (dashed contours) may not generalize well to OOD data (solid contours) in terms of predictive uncertainties. Using the ID classifier boundary (dashed black line), OOD class A predictions are overconfident, while OOD class B ones are underconfident.
    (b) and (c). Logit distributions for the OOD data (b) and ID data (c). See text for details on green star and magenta cross. }
    \label{fig:toy-lr} 
\end{figure}

A potential reason for these referral failures is that the model's predictive uncertainties for ID data do not generalize well to OOD data. Fig. \ref{fig:toy-lr} depicts a simulated example ({Appendix E.2 for details}), illustrating the challenge with naively computing predictive uncertainties for OOD samples with a classifier trained on ID data, especially when ID and OOD data distributions are substantially separated in feature space. Because the ID classification boundary passes close to the mode of OOD class B distribution, even highly typical class B samples (Fig. \ref{fig:toy-lr}a-b, green star) are likely to be referred earlier than samples that lie on the boundary (Fig. \ref{fig:toy-lr}a-b, magenta cross). 

We tested whether domain generalization ~\cite{dg_survey_zhou2022domain} -- learning diagnostic features that enable generalization to OOD images -- would ameliorate these referral failures. Among the many domain generalization approaches for computer vision applications~\cite{li2017deeper}, we chose to evaluate a recently proposed approach for data-efficient generalization that works well for diagnostic medical imaging~\cite{azizi2023}. REMEDIS -- the new framework -- comprises of three stages: i) representation learning - initializing model parameters with supervised pretraining on large-scale, natural image datasets, ii) self-supervised pretraining - adaptation to the medical image domain using contrastive learning on unlabelled images, with the SimCLR algorithm~\cite{simclr_chen2020simple}, iii) fine-tuning - supervised training on labeled ID data. A fourth step -- fine tuning with labeled OOD data -- is optional and not included in our evaluations. Following these steps, the model is tested for its AUROC on an unseen OOD test dataset. 

We tested whether the REMEDIS framework (see section 3.2 for implementation details and key differences with the published method of~\cite{azizi2023}) could rescue these referral failures when applied to the covariate-shifted DR dataset. Here, the model's backbone encoder was initialized with a Vision Transformer (ViT-B/32)~\cite{dosovitskiy2021an} trained for classification on ImageNet21K~\cite{deng2009imagenet}, pretrained with contrastive learning on unlabelled EyePACS and APTOS data, and fine-tuned on the EyePACS data. Following this, we measured the referral performance with AURC. 

Despite a high AUROC (0.95) at zero referral rates, this model also exhibited severe referral failures on the OOD/APTOS dataset, nearly identical to those exhibited by the ResNet-50 baseline model (Fig. \ref{fig:rocs-refs-sec2}b blue). As before, AURC-AUROC was 0.82, considerably worse than the AUROC  at zero referral rates. Similarly, AURC-Accuracy was 0.90, and not much better than accuracy at zero referral rates (0.89). We analyze these failures subsequently (section 3.1) to understand why even robust generalization approaches fail to rescue these severe referral failures.

\subsection{Uncertainty estimation with Bayesian models}
A second potential source of referral failures is the unreliability of the predictive uncertainty measure. It is well known that conventional deterministic models produce significant model miscalibration and often yield overconfident, albeit incorrect predictions, especially when the test data are far removed from the training distribution~\cite{ood_lakshminarayanan2017simple, ood_nalisnick2019detecting}. Label uncertainty, quantified with predictive entropy (equation \ref{eq:entropy}), no longer suffices as a reliable metric for referral.
To address this challenge, \cite{band2021benchmarking} proposed various Bayesian deep learning models that learn distributions over the network parameters. Here, predictive uncertainties consider both aleatoric -- uncertainty due to inherent ambiguity and noise in the data -- and epistemic -- uncertainty due to model constraints or biases in the training process~\cite{gal2016uncertainty}. 

Briefly, the ``total uncertainty'' for Bayesian networks, where the network parameters $\theta$ are stochastic, is given as the sum of its aleatoric and epistemic uncertainties:
\begin{equation}\label{eq:totaluncert}
    \underbrace{\mathcal{H} \left ( \mathbb{E} \left [ p_\theta \left ( y | x \right ) \right ] \right )}_\mathrm{total} 
    = 
    \underbrace{\mathbb{E} \left [ \mathcal{H} \left ( p_\theta \left ( y | x \right ) \right ) \right ]}_\mathrm{aleatoric} + 
    \underbrace{\mathcal{I} (y; \theta)}_\mathrm{epistemic}
\end{equation}
where the expectation is taken over network parameters and $\mathcal{I} (y; \theta)$ represents the mutual information between the outputs and the network parameters. Using this metric \cite{band2021benchmarking} showed consistent improvement over a deterministic baseline, especially for in-domain data, with various Bayesian models including Monte-Carlo dropout (MCD) \cite{gal2016dropout} and mean-field variational inference (MFVI) \cite{xing2002mvfi}.
Yet, surprisingly, even Bayesian uncertainty metrics fail to generalize under covariate shift. To illustrate this, we plot the referral curve based not on predictive entropy (aleatoric uncertainty) alone but based on the total uncertainty. In this case, again, we observed referral failures with AUROC dropping steeply for referral rates $>$10\%. Again, AURC-AUROC (0.86) was 10\% lower than AUROC at zero referral rate (0.96), and AURC-Accuracy (0.93) was only marginally higher than accuracy at zero referral rate (0.89).

In other words, even robust approaches for uncertainty estimation fail on covariate-shifted medical images.
We illustrate this undesirable behavior in the AUROC-referral curve with the MCD model with the ResNet-50 architecture (Fig. \ref{fig:rocs-refs-sec2}b dashed green), among the best performing of the Bayesian deep learning models~\cite{band2021benchmarking}. Nonetheless, a similar pattern of critical failures occurs for the other Bayesian models also, albeit to varying degrees (see Appendix D.1), replicating observations in the original study~\cite{band2021benchmarking} (see {\it their} Figure 5d).

Finally, to realize the ``best of both worlds'', we combined both state-of-the-art approaches: OOD generalization with the REMEDIS framework followed by uncertainty estimation with a Bayesian metric. The latter was computed with Monte-Carlo Dropout in the final fine-tuning step (training and testing stages of step iii) on the ID data (REMEDIS+MCD). Even in this case, sharp referral failures occurred with the AURC-AUROC, and the AURC-Accuracy curve revealed significant non-monotonicity as a function of referral rate (Fig. \ref{fig:rocs-refs-sec2}b-c dashed blue).

\section{Strategies for mitigating referral failures}
\subsection{Diagnosing referral failures}
\label{sec:diagnosing-failures}

\begin{figure}[ht]
    \centering

    \includegraphics[width=0.47\textwidth]{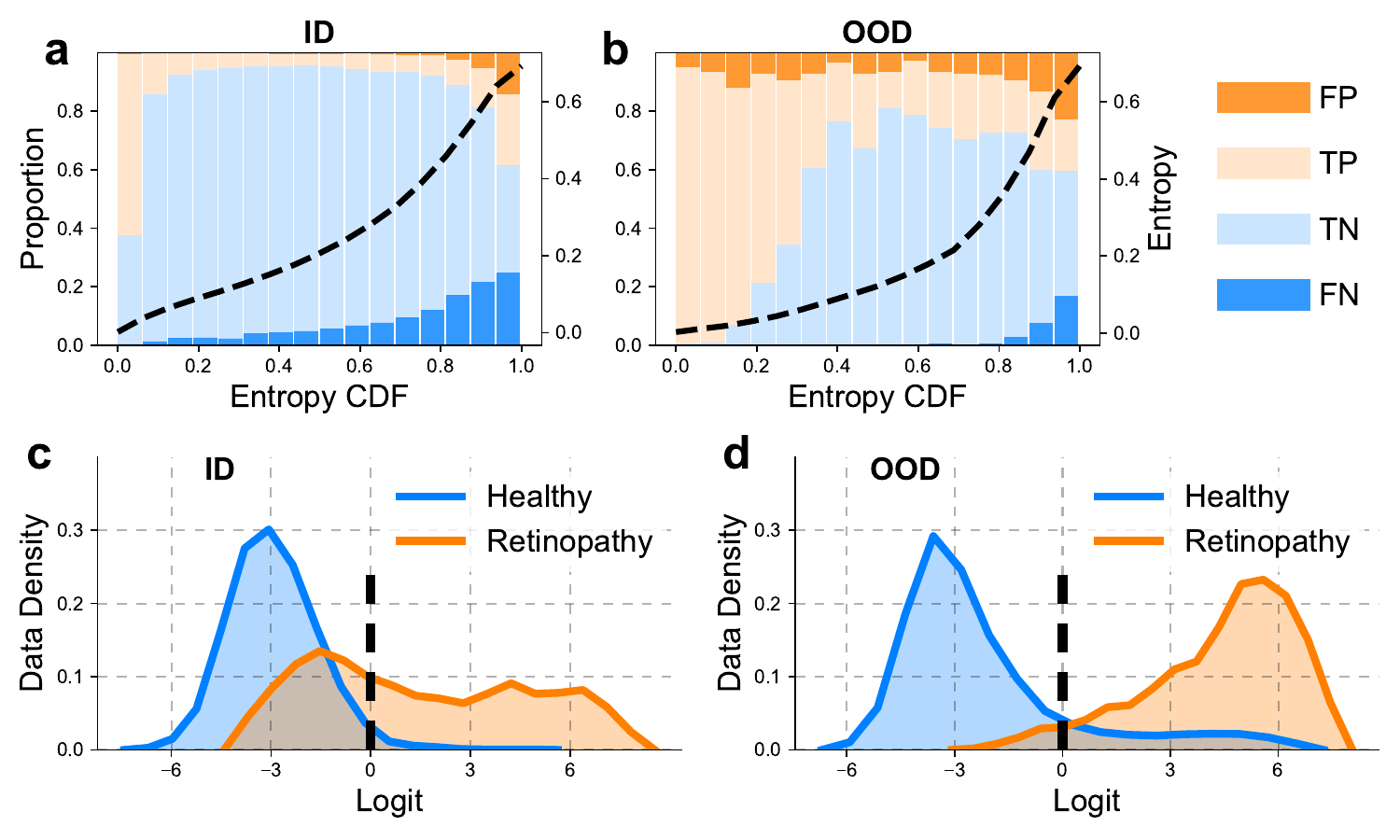}

    \caption{Total entropy and logit distributions for DR data. 
    (a-b) Entropy CDF histograms for ID data (a) and OOD data (b). Light/dark shades are correct/incorrect predictions respectively. True/false negatives (blue), true/false positives (orange). Dashed black line: inverse CDF of entropy.
    (c-d) Logit Distributions for ID data (c) and OOD data (d). x=0 (dashed vertical line) represents the classifier boundary; x$>$0: predicted diseased; x$<$0: predicted healthy. REMEDIS+MCD, with ViT backbone (see text). }

    \label{fig:logits-ech-remmcd}
\end{figure}

We analyze the reasons for these referral failures by visualizing the total entropy distributions produced by the {REMEDIS+MCD} model trained on the EyePACS dataset and tested with the APTOS (Country shift) dataset (Fig.~\ref{fig:logits-ech-remmcd}a-b). For the APTOS test dataset, while true positives and true negatives are predicted with high confidence (low entropy), we find a significant imbalance in entropy among two error types. False negatives are generally predicted with low confidence (high entropy), but a significant fraction of false positives are predicted with high confidence (low entropy). 

To understand the reason behind this trend, first, we visualize the distribution of the model outputs (logits, $f(x)$) (Fig.~\ref{fig:logits-ech-remmcd}c-d). The dashed vertical line in the figure denotes the model's classification boundary estimated from the ID data, such that positive (diseased) and negative (healthy) label assignments are made to either side of this boundary. Note that there exists a monotonic relationship between the total entropy ($\mathcal{H}$) and the logit magnitude: for any referral rate based on an entropy threshold ($\mathcal{H}_r$), there is a unique threshold $L_r$ on the logit magnitude, such that all the test samples with $f(x) \in [-L_r, L_r]$ are referred.

From the visualization, it is apparent that healthy class OOD distribution has a long tail to the right of the classification boundary (false positives, Fig.~\ref{fig:logits-ech-remmcd}d, blue distribution to the right of the classification boundary, Logit = 0). Thus, some of these samples would be predicted, albeit incorrectly, with even greater confidence than the most confident true negatives! As a result, a significant proportion of the most confident predictions are also incorrect. These incorrectly labeled samples are retained with high confidence at higher referral rates, leading to a precipitous drop in AUROC. Nearly identical trends were obtained for total entropy and logit distributions estimated with Monte Carlo dropout on a ResNet-50 model ({Appendix E.6}).

\begin{figure}[b]
    \centering
 
    \includegraphics[width=0.47\textwidth]{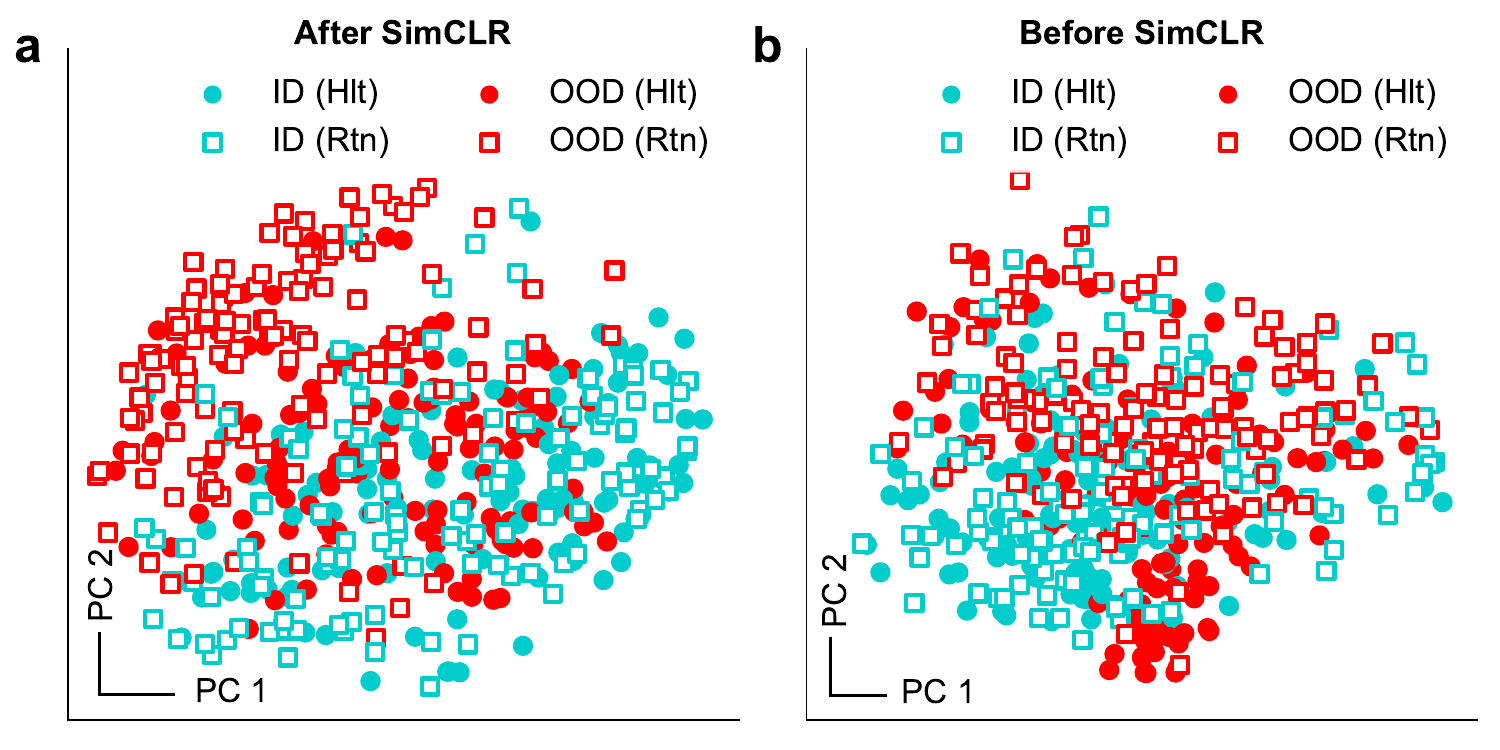}

    \caption{Visualization of ViT encoder representations (top 2 principal components) of DR ID and OOD samples during various stages of REMEDIS. 
    (a) After self-supervised, contrastive learning (after SimCLR/step ii)
    (b) Before contrastive learning (after ImageNet21k pretraining/step i) }

    \label{fig:pca}
\end{figure}

To further understand these failures, we visualize the logit representations themselves using principal component analysis (PCA) in two dimensions. We find that the SimCLR objective segregates both classes of ID from both classes of OOD representations, such that the inter-domain separation (ID versus OOD) is more pronounced than the inter-class separation (healthy versus diseased) (Fig. \ref{fig:pca}a). In fact, ID and OOD representations are even more segregated following self-supervised pretraining (step ii) than before this step, i.e., with the initial ViT model (step i) (Fig. \ref{fig:pca}b). We quantify the segregation at each step in {Appendix E.3}.

We expect that this segregation is a direct consequence of the contrastive loss in the self-supervised training objective: because the SimCLR algorithm is agnostic to the specific diagnostic features that are relevant for classification, low-level image features that significantly differ between domains (e.g., overall contrast, image resolution) may drive the clustering of OOD samples away from ID samples.

\subsection{Robust generalization approaches}
To overcome these out-of-domain referral failures, we explore novel classes of generalization approaches. Each approach falls within the REMEDIS framework, and several employ novel combinations of self-supervised learning objectives. In all cases, the model is a Vision Transformer (ViT) architecture initialized by training with a classification objective for ImageNet21K. In the first two cases, the model is fine-tuned with the respective ID medical images~\cite{azizi2023} sequentially, in the final stage. The set of approaches is summarized in Fig. \ref{fig:schematic}d.

\textbf{Contrastive Learning (SimCLR)}:
The SimCLR~\cite{simclr_chen2020simple} objective seeks to maximize the cosine similarity between representations of the same image under different augmentations (e.g. random cropping, random saturation), using a contrastive loss. A detailed description is provided in {Appendix A.3}. We call this approach REMEDIS/SimCLR (Fig. \ref{fig:schematic}d).
The study that proposed the REMEDIS framework tested a variety of contrastive loss functions (MoCo \cite{chen2020improved}, RELIC \cite{mitrovic2021representation}, Barlow Twins \cite{zbontar21a}), and found them to be comparable to SimCLR in their efficacy. Yet, as discussed previously (section \ref{sec:diagnosing-failures}), a contrastive loss objective may cause OOD samples to be clustered away from the ID samples. Therefore, we evaluate two other self-supervised objectives, neither of which involves a contrastive loss.

\textbf{Masked Image Modeling (SimMIM) and Masked AutoEncoder (MAE)}:
In the SimMIM approach~\cite{simmim_xie2022simmim}, each image is divided into $P$ patches. A subset of patches are randomly masked and passed on to the ViT encoder. The resultant embeddings are passed to the ``prediction head'' which predicts the pixels of the masked patches. We call this approach REMEDIS/SimMIM (Fig. \ref{fig:schematic}d).

The MAE approach~\cite{he2022masked}, is closely similar to the SimMIM approach except that only unmasked patched as passed to the ViT encoder, which generates their respective embeddings. Then, these are passed along with the masked patches to the decoder which predicts the pixels of the masked patches. Because only the unmasked patches are passed to the encoder, compute and memory requirements are lower. We call this approach REMEDIS/MAE (Fig. \ref{fig:schematic}d). The optimization objective for both SimMIM and MAE is described in {Appendix A.4}.

Both SimMIM and MAE enable capturing local, high spatial frequency features, such as texture, whereas contrastive learning favors capturing global, low spatial frequency features \cite{park2023what}. Medical images in which the symptoms are localized anomalies (e.g. focal damage to retinal blood vessels) could benefit from SimMIM and MAE.



\textbf{Domain Adversarial Network (DAN)}:
Domain adversarial training of neural networks~\cite{dan_ganin2016domain} concurrently learns efficient representations for ID and OOD data, optimizing both a label loss for ID samples and a domain loss that penalizes representations that distinguish ID from OOD data, using a ``gradient reversal'' layer. Like SimCLR and SimMIM/MAE, OOD image labels are not used in training, but unlike the other approaches, the fine-tuning of the model with ID images occurs concurrently with  domain adaptation. The optimization objective is defined in {Appendix A.6}. We call this approach REMEDIS/DAN (Fig. \ref{fig:schematic}d).
This approach encourages the model to ignore domain-distinctive features, i.e., those exhibiting high dissimilarity across domains. As a result, when class-distinctive features are independent of domain-distinctive features, the model can rely on the latter alone for label predictions. A simulated example is provided in Section \ref{sec:dan-effectiveness}.

\begin{figure*}[t]
    \centering
    \includegraphics[width=\textwidth]{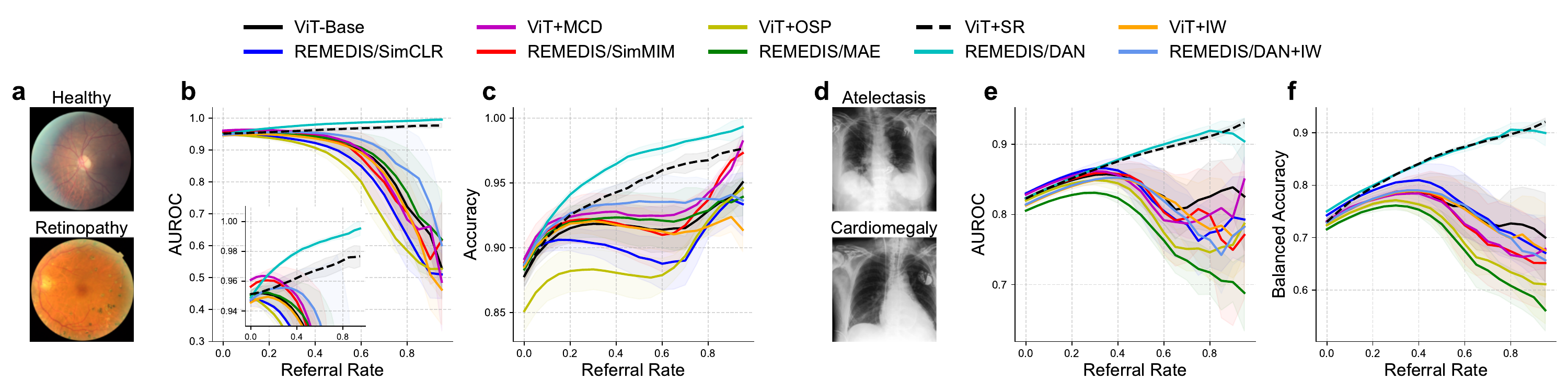}

    \caption{
        OOD referral performance in two medical domains.
        (a) Sample diabetic retinopathy (DR) images
        (b-c) DR AUROC (b) and accuracy (c) referral curves. 
        (d-f) Same as in (a-c), but for the chest X-ray (CX) dataset, averaged across all 5 conditions.
    }
    \label{fig:ref-curves-main}
\end{figure*}

\textbf{Importance Weighting (IW)}:
Importance weighting is a domain adaptation approach where each training (ID) sample contributes differently to the total classification loss, according to some measure of its proximity to the OOD distribution. For a target OOD distribution, the weight for a given sample $x$ is specified as the ratio of OOD density to ID density at that $x$ \cite{shimodaira2000iw}.
The ratio may be estimated using a domain discriminator \cite{bickel2007iw}: $\frac{p_\text{\tiny OOD}(x)}{p_\text{\tiny ID}(x)} \approx \frac{g(x)}{1-g(x)}$, where $g$ is a binary classifier whose positives and negatives correspond to OOD and ID respectively. In this way, the classification boundary for the ID data is influenced more by samples that overlap with the OOD distribution, than otherwise. We combine the pretrained ViT backbone from REMEDIS (step i) with importance weighting (step iii); we call this approach ViT+IW. 

\cite{park2020calibratediw} proposed combining domain adversarial feature learning with importance weighting and showed improvements in calibration for certain digit datasets. Inspired this idea, we combined the pretrained ViT backbone from REMEDIS (step i) with DAN (step ii) and IW (step iii), the last two being concurrent. We call this approach REMEDIS/DAN+IW. Finally, we also evaluate a baseline model that does not include self-supervised pretraining or domain adaptation, which we refer to as ViT-Base (Fig. \ref{fig:schematic}d).

\subsection{Robust referral strategies}
Next, we explore novel combinations of recently proposed referral strategies (MCD, OSP, SR) for robust selective classification with the REMEDIS ViT backbone. In each case, the ViT backbone was initialized on ImageNet 21k (step i), followed by fine-tuning (step iii) but without the self-supervision step (step ii) (Fig. \ref{fig:schematic}d).

\textbf{Monte Carlo dropout (MCD)}:
Recent work suggests that Bayesian neural networks that take into account the total uncertainty of label predictions (\ref{eq:totaluncert}) enable more reliable selective classification~\cite{band2021benchmarking}. Yet, because learning a distribution over network parameters is computationally expensive, \cite{gal2016dropout} proposed using Monte Carlo dropout as an effective approximation for variational inference in such neural networks. Mathematically, a deterministic network is trained by solving the optimization problem: $\arg \min _{\boldsymbol{\theta}}\left\{-\mathbb{E}_q\left[\log p\left(\mathbf{y}_{\mathcal{D}} \mid f(\mathbf{X} ; \boldsymbol{\theta})\right)\right]+\lambda\|\boldsymbol{\theta}\|_2^2\right\}$, where $q_{\Theta}$ is the distribution over network parameters obtained with dropout. We implement MCD on the base ViT architecture, specified by the hyperparameter of dropout rate, applied to hidden layer nodes and attention softmax weights. We call this approach ViT+MCD (Fig. \ref{fig:schematic}d).

\textbf{One-sided Prediction (OSP)}:
\cite{osp_gangrade2021selective} proposed a selective classification technique that modifies the loss function by introducing a separate classifier for each label. For instance, for data with $K$ labels, $K$ one-vs-all binary classifiers are trained  -- $f_{1}(\cdot)$, $f_{2}(\cdot)$, ..., $f_{K}(\cdot)$ -- where $f_{i}(x)$ reflects the predicted probability that $x$ belongs to class $i$ ($p(y=i|x)$). Because $K$ different binary classifiers are trained, these prediction probabilities will generally not sum to 1. The final predicted label is $k^* = \text{argmax}_k f_k(x)$ and uncertainty is inverse of $f_{k^*}(x)$. If the minimum prediction probability (corresponding to the predicted label) is below a preset threshold $\tau$ (determined by the referral rate), then the model abstains from prediction, and the sample is referred (loss function described in {Appendix A.5}). We call this approach ViT+OSP (Fig. \ref{fig:schematic}d).

\textbf{Split Referral (SR)}:
\cite{anuj_srivastava2023avoiding} proposed an approach for robust referral by eliminating uncertainty imbalance across classes. First, OOD test samples are ranked separately for each predicted class based on their entropies (aleatoric or total, see Experiments). Then, referrals are made in the order of the least to the most confident predictions proportionately for each predicted class.
Mathematically, let $\mathcal{N} = \{x \in \mathcal{D}_\mathrm{test} : f_\theta(x) < 0.5\}$, and $\mathcal{P} = \{x \in \mathcal{D}_\mathrm{test} : f_\theta(x) \ge 0.5\}$ be the sets of negative and positive predictions respectively among the test set $\mathcal{D}_\mathrm{test}$; where the model $f_\theta(\cdot)$ which predicts the probability of a positive label for a given input. For a given referral rate $r$, $r |\mathcal{N}|$ and $r |\mathcal{P}|$ samples with the highest entropy are referred from $\mathcal{N}$ and $\mathcal{P}$ respectively, and the proportion of predicted labels is maintained at every rate of referral. We call this approach ViT+SR (Fig. \ref{fig:schematic}d).

\section{Experiments}

\subsection{Datasets}
We employ two different datasets -- one for diabetic retinopathy (DR) prediction with retinal fundus images and another for multilabel prediction with chest X-ray (CX) images. For Diabetic Retinopathy, datasets, and labeling conventions follow an approach identical to previous work~\cite{band2021benchmarking}. Briefly, the model seeks to classify retinal fundus images into one of two classes (binary classification) based on the severity of DR (Fig.~\ref{fig:ref-curves-main}a); details are provided in {Appendix C.1}. For Chest X-ray datasets -- ID: CheXpert \cite{irvin2019chexpert} and OOD: ChestX-ray14 \cite{Wang_2017_CVPR} -- and labeling conventions we follow previous work~\cite{azizi2023}, except for a few minor differences (see {Appendix C.2}). Briefly, the objective is multilabel  classification into one of 5 conditions (Fig.~\ref{fig:ref-curves-main}d; {Appendix C.2}). 



\subsection{DR referral performance}
We compared the performance of each of the models against the baseline model (ViT-Base). All methods except ViT+SR and REMEDIS/DAN suffered non-monotonicity in their referral curves for DR (Fig. \ref{fig:ref-curves-main}b-c). ViT+SR achieved AURCs of 0.94 (Accuracy) and 0.96 (AUROC), while REMEDIS/DAN was the best performer with AURCs of 0.96 (Accuracy) and 0.98 (AUROC). Overall, REMEDIS/DAN achieved gains of 4.7\% in AURC-accuracy and 12.8\% in AURC-AUROC over ViT-Base, while ViT+SR gained 2.9\% in AURC-accuracy and 11.1\% in AURC-AUROC over ViT-Base. 

\begin{table}[t]
\centering
\begin{tabular}{lcccc}
\toprule 
                                  & \multicolumn{2}{c}{\textbf{Retinopathy}}                                                                                                & \multicolumn{2}{c}{\textbf{Chest X-ray}}                                                                                                     \\
                                  
\multirow{-2}{*}{\textbf{Method}} & \textbf{\begin{tabular}[c]{@{}c@{}}AURC\\ AUROC\end{tabular}} & \textbf{\begin{tabular}[c]{@{}c@{}}AURC\\ Acc.\end{tabular}} & \textbf{\begin{tabular}[c]{@{}c@{}}AURC\\ AUROC\end{tabular}} & \textbf{\begin{tabular}[c]{@{}c@{}}AURC\\ B Acc.\end{tabular}} \\ \midrule
ViT-Base                          & 0.8675                                                         & 0.9172                                                        & 0.8363                                                         & 0.7537                                                             \\
ViT+MCD         & 0.8699                                                         & 0.9296                                                            & 0.8272                                                         & 0.7354                                                                       \\
\bl{ViT+OSP}         & 0.8052                                                         & 0.8905                                                            & 0.8042                                                         & 0.7098                                                                       \\
\textbf{\textit{\bl{ViT+SR}}} & {{\textit{0.9642}}}                         & {{\textit{0.9441}}}                            & {{\textit{0.8799}}}                         & {{\textit{0.8443}}}                                       \\
R/SimCLR        & 0.8265                                                         & 0.9041                                                            & 0.8250                                                         & 0.7603                                                                       \\
\bl{R/SimMIM}        & 0.8575                                                         & 0.9235                                                            & 0.8225                                                         & 0.7356                                                                       \\
\bl{R/MAE}           & 0.8789                                                         & 0.9214                                                            & 0.7804                                                         & 0.6939                                                                       \\
\textbf{\bl{R/DAN}} & {{{\textbf{0.9786}}}}                         & {{\textbf{0.9603}}}                            & {{\textbf{0.8817}}}                         & {{\textbf{0.8463}}}                             \\
\bl{ViT+IW}          & 0.8513                                                         & 0.9145                                                            & 0.8223                                                         & 0.7614                                                                       \\
\bl{R/DAN+IW}        & 0.9068                                                         & 0.9297                                                            & 0.8173                                                         & 0.7461                                                                       \\
\bottomrule
\end{tabular}
\caption{
`R'- REMEDIS and `B Acc.' - Balanced Accuracy. 
\textbf{Bold} and \textit{italics} denotes the highest and second-highest AURC, respectively.
Blue: novel combinations evaluated. For DR and CX, all AURCs were significantly higher for ViT+SR and REMEDIS/DAN than other methods (two-sided \textit{t}-test, p$<$0.05). 
}
\label{table:aurc}
\end{table}

\subsection{CX referral performance}
Chest X-ray referral is arguably more challenging than DR referral because  target labels for the ID (train split) and OOD datasets are extracted with NLP and, consequently, noisy. Also, the label extraction method differed between ID and OOD, leading to behavior drift on top of covariate shift. Additionally, the proportion of positive labels in OOD data (2-12\%) is significantly lower than in the training data (46-59\%). Here, we consider only proportion-agnostic metrics (e.g., balanced accuracy or AUROC); addressing class imbalance is scope for future work. A sample was considered negative only if all 5 disease categories were absent.

All methods except ViT+SR and REMEDIS/DAN suffered non-monotonicity in their referral curves, on average (across all 5 diseases) (Fig. \ref{fig:ref-curves-main}e-f). Both ViT+SR and REMEDIS/DAN achieved closely similar AURCs of 0.84 (Balanced Accuracy) and 0.88 (AUROC), gaining {12.3\% and 5.4\%} respectively over ViT-Base (Table \ref{table:aurc}).
We also analyzed AURC on individual classes of CX diseases to see how the models performed for individual conditions; the full set of results is reported in Appendix E.4. Overall, non-monotonicity of the referral curve was evident for 3 of the 5 conditions. Across conditions, we observed improvements up to 26\% and 15\% in AURC-Balanced Accuracy and AURC-AUROC respectively, over ViT-Base.

To rule out concurrent optimization as the reason for REMEDIS/DAN's better performance, we also sought to train a model by simultaneously minimizing the SimCLR and classification losses. This model (REMEDIS/SimCLR-joint) also exhibited marked referral failures {Appendix E.5}.

\subsection{Analysis}
\label{sec:dan-effectiveness}

\begin{figure}[ht]
    \centering
    \includegraphics[width=0.47\textwidth]{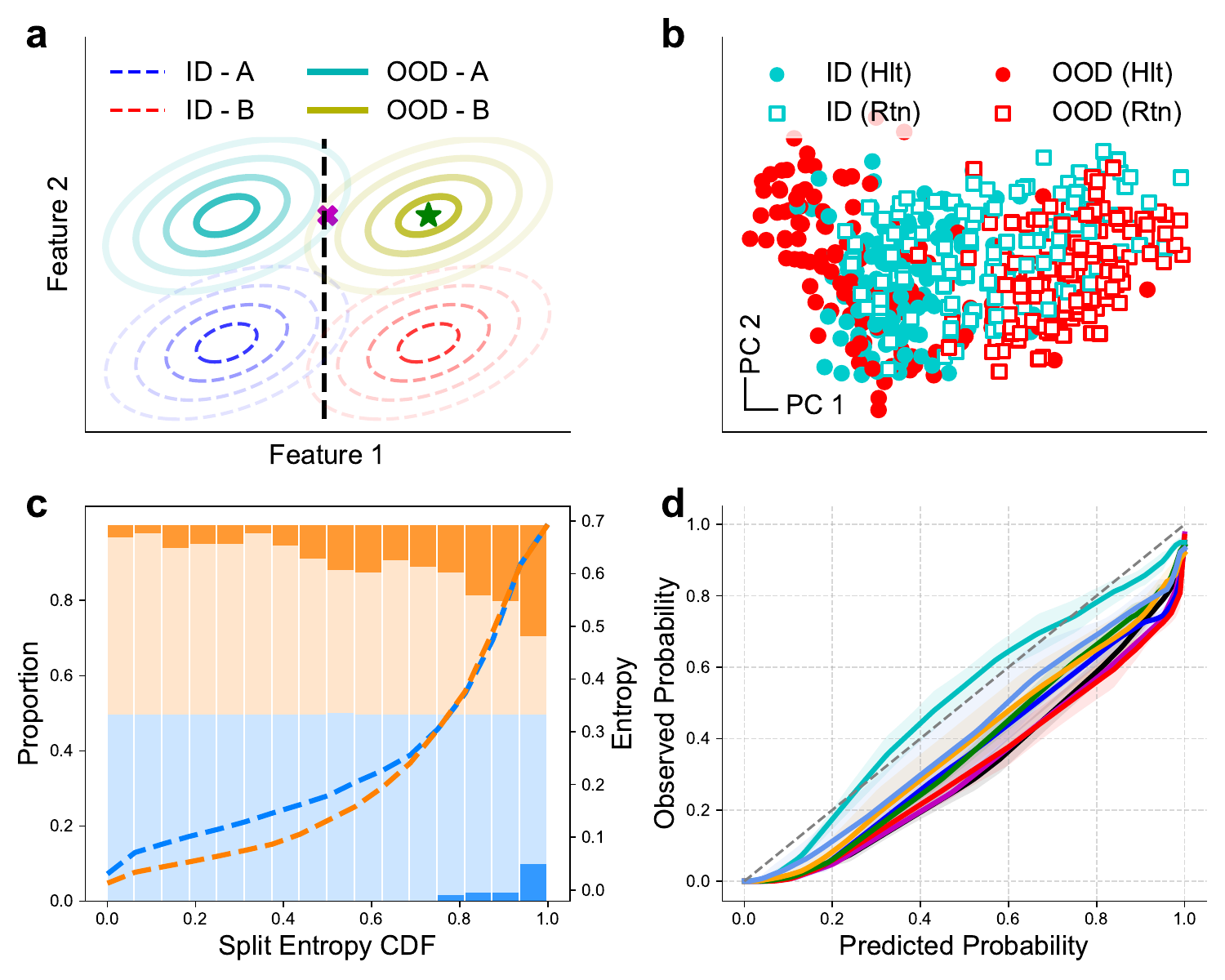}
    \caption{Explaining DAN and SR success. (a) Domain adversarial logistic regression (dashed black line) on simulated data; same as in Fig. 3a. (b) Visualizing representation of DR ID and OOD samples after DAN training; other conventions as in Fig. \ref{fig:pca}. (c) Entropy CDF histogram on DR OOD data with ViT+SR. Dashed blue and dashed orange curves: inverse CDF within healthy and diseased prediction sets respectively; other conventions as in Fig. \ref{fig:logits-ech-remmcd}a-b. (d) Calibration curves on DR OOD data; legend in Fig. \ref{fig:ref-curves-main}.}
    \label{fig:dan-eff}
\end{figure}

We analyze the reasons for the effectiveness of REMEDIS/DAN and ViT+SR. First, we analyze REMEDIS/DAN with the same simulated example as shown in Fig. \ref{fig:toy-lr}. Unlike a classifier trained on ID data alone, a classifier using a domain adversarial feature extractor can access  unlabelled OOD data (Fig. \ref{fig:dan-eff}a). This classifier, therefore, learns the domain invariant feature (feature 1) while ignoring the domain distinctive feature (feature 2), thereby enabling successful OOD generalization. DAN also overcomes the challenge with ID and OOD segregation that occurred with SimCLR (Fig. \ref{fig:dan-eff}b).

Next, we investigated the effect on model calibration, a desirable property for reliable referral \cite{calibrationreferral}. We quantified miscalibration with the expected calibration error (ECE) that measures the expected deviation of the predicted probability from the observed probability of samples belonging to the positive class. We observed systematic improvements in ECE with REMEDIS/DAN, ranging from 14\% to 60\% across all conditions (Table \ref{table:ECE}, see also Appendix E.4). In contrast, for self-supervision and Bayesian methods, ECEs did not differ significantly from the base model; these models could not guarantee calibration under covariate shift. 

Yet, reducing miscalibration is not the only solution. In Section \ref{sec:motivn}, we showed that referral failure occurs (i) when model uncertainties are imbalanced across classes and (ii) larger error rates occur in low uncertainty regions (miscalibration); REMEDIS/DAN improves referral by addressing the latter.  In {Appendix D}, we show that as long as the calibration curve increases monotonically (Fig. \ref{fig:dan-eff}d), SR yields a monotonically increasing accuracy-referral curve, regardless of calibration error. In other words, ViT+SR rescues referral failures by eliminating uncertainty imbalance.

Domain adaptation methods are inherently designed to handle covariate shifts. Yet, not all of these methods successfully addressed referral failures under covariate shift. Of the methods tested, REMEDIS/DAN significantly outperformed other methods. Surprisingly, methods incorporating importance weighting (ViT+IW and REMEDIS/DAN+IW) did not. Training with IW is known to be challenging for over-parameterized deep networks ~\cite{what2019-iw, xu2021understanding-iw, rethink2020-iw}. In Appendix  E.8, we show additional experiments where IW improves OOD referral at the cost of ID performance.


\begin{table}[t]
\centering
\begin{tabular}{lcc}
\toprule
\multirow{1}{*}{\textbf{Method}} & \textbf{Retinopathy} & \textbf{Chest X-Ray} \\
\midrule
ViT-Base & 0.1041 & 0.2047 \\
ViT+MCD & 0.1027 & 0.2086 \\
REMEDIS/SimCLR & 0.1009 & 0.1922 \\
\bl{REMEDIS/SimMIM} & 0.1091 & 0.2135 \\
\bl{REMEDIS/MAE} & 0.0879 & 0.2191 \\
\textbf{\bl{REMEDIS/DAN}} & {\textbf{0.0390}} & {\textbf{0.1521}} \\
\bl{ViT+IW}& 0.0888 & 0.2014 \\
\bl{REMEDIS/DAN+IW} & 0.0784 & 0.1907 \\
\bottomrule                       
\end{tabular}
\caption{Expected Calibration Error. For Chest X-ray, ECE is averaged over  5 conditions. Blue: novel combinations. ViT+SR does not change the outputs of the baseline model; its ECE is the same as ViT-Base. ViT+OSP gives individual probabilities for each class; its ECE is not comparable with other models. REMEDIS/DAN achieves the lowest ECE (p$<$0.05, \textit{t}-test).
}
\label{table:ECE}
\end{table}

\section{Discussion}
Deep learning models have the potential to revolutionize diagnostic medical imaging; yet, referral to a clinician becomes necessary when label predictions are not confident~\cite{band2021benchmarking}. We show that even recent domain generalization methods~\cite{azizi2023} yield miscalibrated predictions on OOD data, leading to severe referral failures. We evaluate several novel strategies to overcome these failures. Overall, our results suggest that such approaches must be evaluated with data from the appropriate demographic in settings in which these models are most likely to be utilized, before deployment.


\phantomsection
\bibliographystyle{unsrt}
\bibliography{main}


\clearpage
\section*{A. Model Specifications}

\subsection*{A.1. Backbone Architectures}
The baseline DR models (R50 and R50+MCD) in Section \ref{sec:motivn} have a ResNet50 backbone with random initialization. Further details are provided in \cite{band2021benchmarking}.

For all other models and experiments we use the JAX / Flax framework, following the 
{uncertainty-baselines}\footnote{\url{https://github.com/google/uncertainty-baselines}} benchmark.
All models have ViT-B/32 as the backbone encoder, initialized with ImageNet21k classification checkpoint\footnote{\url{https://storage.googleapis.com/vit_models/augreg/B_32-i21k-300ep-lr_0.001-aug_light1-wd_0.1-do_0.0-sd_0.0.npz}}.
The encoder hyperparameters are as follows:
\begin{itemize}
    \item patch size: 32 $\times$ 32
    \item hidden size: 768
    \item MLP dimension: 3072
    \item number of attention heads: 12
    \item num of transformer layers: 12
    \item classifier type: auxiliary CLS token
\end{itemize}

The following subsections describe additional components of each model.

\subsection*{A.2. Monte Carlo Dropout}
We implement MCD on the ViT backbone, with dropout applied to all hidden units and attention softmax weights of the transformer blocks. The dropout rate is a hyperparameter. With this model, predictive uncertainty is specified by the total entropy of the average prediction probability, which is computed as follows: dropout is applied on the trained network during inference, and average prediction probability is computed by averaging the sigmoid outputs from 5 Monte Carlo samples of the network parameters, as specified by \cite{band2021benchmarking}.

\subsection*{A.3. SimCLR}
For self-supervised learning with SimCLR, a minibatch of K images is randomly sampled from the unlabelled dataset. Then, 2K augmentations are derived by generating two different augmentations per image. The two augmentations $x_{i}$ and $x_{j}$ from each image (positive pairs) are passed through the pre-trained ViT encoder ($f_{\theta}(\cdot)$) yielding representations $\boldsymbol{z_{i}}$ and $\boldsymbol {z_{j}}$. The contrastive loss between each positive pair of samples $i$, $j$ is computed as:  
\begin{equation} \label{eqn:simclr_loss}
\ell_{i, j} 
=
-\log \frac{\exp \left(\operatorname{sim}\left(\boldsymbol{z_i}, \boldsymbol{z_j}\right) / \tau\right)}{\sum_{k=1}^{2 K} \mathbb{I}[k \neq i] \exp \left(\operatorname{sim}\left(\boldsymbol{z_i}, \boldsymbol{z_k}\right) / \tau\right)}
\end{equation}
where $sim(\cdot, \cdot)$ measures the similarity between two vectors, and $\tau$ is a temperature parameter that determines the permissivity of the similarity measure. The final SimCLR loss is computed as the sum of the losses between all positive pairs in the mini-batch. 
We implement REMEDIS/SimCLR with a 2-layer MLP projection head on the penultimate (pre-logit) embedding of ViT's CLS token, and the output of this projection head is used for the contrastive loss in equation (\ref{eqn:simclr_loss}). We consider the minimum cropping area as a hyperparameter.

\subsection*{A.4. SimMIM and MAE}
Each image is divided into $P$ patches of size $m \times n$, and a subset ($k\%$) of patches is randomly masked.
In the SimMIM approach, the pixel values in the masked patches are set to zero, and the resultant image is passed to the encoder. From the encoder representations, a 2-layer MLP decoder predicts the image pixels.
In the MAE approach, the masked patches are removed from the input, i.e. only the unmasked patches are passed as tokens to the encoder. Tokens consisting of a learnt embedding are introduced to the encoder output at the masked locations. A single-layer transformer decoder then predicts image pixels.
The loss function for the $p^{th}$ masked patch is as:  
\begin{equation}
    \ell_{p} 
    =
    \frac {1}{m n} \sum_{i=1}^{m n}y_i^p \log \frac{y_i^p}{\hat{y}_i^p} + (1-y_i^p) \log \frac{1-y_i^p}{1-\hat{y}_i^p}
\end{equation}
where ${y_i^p}$ and ${\hat{y}_i^p}$ represent the actual and predicted values (0 to 1) of the $i^{th}$ pixel of the $p^{th}$ patch.  The final loss is computed as the summation of the losses of all masked patches. 
For both approaches, the masking rate (probability with which a patch is masked) is a hyperparameter.

\subsection*{A.5. One-Sided Prediction}
For one-sided prediction\cite{osp_gangrade2021selective}, the optimization objective for the one-vs-all classifier $f_k(\cdot; \theta)$ corresponding to the $k^{th}$ class is as follows (simplified from the original study):
\begin{equation} \label{eqn:osp_loss}
\begin{split}
\min_{\theta, \varphi}\ \max_{\lambda \ge 0}\ & \frac{1}{n_k} \sum_{i:y_i=k} -\log f_k(x; \theta) + \varphi_k\\
+ & \lambda_k \left( \frac{1}{n_{\neq k}} \sum_{i:y_i \neq k} -\log (1 - f_k(x; \theta)) - \varphi_k  \right)
\end{split}
\end{equation}
Extending the base ViT network, the ViT+OSP model includes an additional head for computing the loss in equation (\ref{eqn:osp_loss}) with Lagrangian multipliers and error threshold parameters. The hyperparameters are the learning rate ratios for updating $\lambda$s and $\varphi$s, relative to the base learning rate for the network.

\subsection*{A.6. Domain adversarial training}
For the DAN model, the optimization objective is defined as follows:
\begin{equation}
\begin{split}
    & \min_{C, E}\ \max_{D}\ {\mathbb{E}}_\text{\tiny ID}[\ell(y, C \circ E(x))] \\
    & + {\mathbb{E}}_\text{\tiny ID}[\log(1 - D \circ E(x))] + {\mathbb{E}}_\text{\tiny OOD}[\log D \circ E(x)]
\end{split}
\end{equation}
where $E$ is the backbone encoder (in this case, ViT), $C$ is the classifier head (single layer), $D$ is the domain predictor (MLP) whose output range is $[0,1]$, $\ell$ is the classification loss (cross entropy), $x$ and $y$ are input sample and target label respectively. $D$ minimizes a binary cross entropy loss where positive samples belong to the OOD dataset and negative samples belong to the ID dataset. $E$ is trained to maximize the domain prediction loss, which results in extracting domain invariant features from the raw inputs.

We implement the domain predictor as an MLP on the penultimate (pre-logit) embedding of the CLS token. The minimax objective is implemented as gradient reversal layer at the encoder-domain predictor interface, i.e. gradients are reversed between the domain predictor input and encoder output during backpropagation. The hyperparameters here are the depth and hidden dimension (width) of the domain predictor MLP, and the gradient multiplier at the reversal layer (equivalent to ratio of learning rates of the max and min objectives).

\subsection*{A.7. Importance Weighting}
For IW, the architecure is the same as in DAN.
The objective for the domain predictor is to minimize:
\[
{\mathbb{E}}_\text{\tiny ID}[-\log(1 - D \circ E(x))] + {\mathbb{E}}_\text{\tiny OOD}[-\log D \circ E(x)]
\]
where $E$ is the backbone encoder, $D$ is the domain predictor (MLP) whose output range is $[0,1]$, $x$ is the input sample.
The objective for the classifier and encoder is to minimize:
\[
{\mathbb{E}}_\text{\tiny ID} \left[ \tfrac{D \circ E(x)}{1 - D \circ E(x)} \cdot \ell(y, C \circ E(x)) \right]
\]
where $C$ is the classifier head (single layer), $y$ is the target label and $\ell$ is the classification loss (cross entropy).
For ViT+IW, the above objectives are minimized concurrently but independently, by stopping the gradients between $D$ and $E$.
For REMEDIS/DAN+IW, we use the gradient reversal implementation as described in Appendix A.6.

\subsection*{A.8. General Training Details}
Hyperparameter search spaces and final values for each model are provided in Table \ref{table:a1}.
All models are optimized using SGD with momentum ($\beta = 0.9$) having a base learning rate of $10^{-3}$ with linear decay, and gradient clipping at norm $1.0$.
The source codebase includes all model architecture definitions, hyperparameter configuration files and training scripts.

All reported scores are averaged over 6 runs, each with a different random seed. We use JAX's pseudorandom number generator for all executions, except random augmentations/masking for SimCLR/SimMIM/MAE pretraining, which were generated with \texttt{tensorflow.image} and \texttt{tensorflow.keras.backend} modules. Thus the SimCLR, SimMIM MAE models may produce slightly different results with each run; therefore, averaging over the 6 random seeds improves the reproducibility of the performance numbers.

We used one NVIDIA A100 Tensor Core GPU (40GB RAM) for all training. OS is Ubuntu 20.04. All software requirements with version details and installation instructions would be included in the codebase.

\begin{table}[t]
\centering
\begin{tabular}{ccc}
\toprule
\textbf{Method} & \textbf{Hyperparameter} & \textbf{Search Space} \\
\midrule
MCD & dropout rate & 0.01$^b$, 0.03$^a$, 0.1, 0.3 \\
\midrule
\multirow{2}{*}{OSP} & relative lr ($\lambda$) & 0.1$^b$, 1, 10$^a$ \\
 & relative lr ($\varphi$) & 0.3, 1$^b$, 3$^a$ \\
\midrule
SimMIM & masking rate & 0.1, 0.2, 0.4$^b$, 0.6$^a$, 0.8 \\
\midrule
MAE & masking rate & 0.1$^b$, 0.2, 0.25$^b$, 0.4, 0.6, 0.8 \\
\midrule
SimCLR & min crop area & 0.36, 0.64$^{ab}$, 0.81\\
\midrule
\multirow{3}{*}{DAN} & grad multiplier & 0.1, 0.3$^b$, 1$^a$, 3, 10 \\
 & dom-pred depth & 2$^a$, 3, 4, 5$^b$ \\
 & dom-pred width & 256$^a$, 512$^b$, 768 \\
\midrule
IW & domain loss coeff & 3$^{ab}$, 10\\
\midrule
\multirow{2}{*}{DAN+IW} & domain loss coeff & 3, 10$^{ab}$ \\
 & grad multiplier & 0.1, 1$^{ab}$, 10 \\
\bottomrule
\end{tabular}
\caption{
Hyperparameters tested, and values selected, for Diabetic Retinopathy (superscript$^a$) and Chest X-ray Conditions (superscript$^b$) based on maximum AUROC for the ID validation split.
}
\label{table:a1}
\end{table}

\section*{B. Evaluation Metrics}
To measure classification performance, we use the area under the receiver operating characteristic curve (AUROC) for Retinopathy and all Chest X-ray conditions, accuracy (Acc) for Retinopathy, and balanced accuracy (bAcc) for all Chest X-ray conditions. The latter metric is relevant to address extreme class imbalances: the proportion of positive samples across ID and OOD varied considerably across datasets (46-59\% for ID and 2-12\% for OOD).
\begin{equation}
    \text{AUROC}(\mathcal{D}) = \frac{1}{n_0 n_1} \sum_{i:y_i=1} \sum_{j:y_j=0} \mathbb{I}[f_\theta(x_i) > f_\theta(x_j)]
\end{equation}
\begin{equation}
    \text{Acc}(\mathcal{D}) = \frac{1}{n} \sum_{i} \mathbb{I} [y_i = \text{round}(f_\theta(x_i))]
\end{equation}
\begin{equation}
    \text{bAcc}(\mathcal{D}) = \frac{1}{2} \sum_{k \in \{0,1\}} \frac{1}{n_k} \sum_{i:y_i=k} \mathbb{I} [y_i = \text{round}(f_\theta(x_i))]
\end{equation}
where $f_\theta(\cdot)$ is the model's prediction probability for class 1, $n_0$ \& $n_1$ are number of negative and positive points respectively, and $\mathcal{D} = \{(x_i, y_i)\}_{i=1}^n$.
These are computed with the \texttt{sklearn.metrics} module.

Referral-based selective classification performance is measured with the area under referral curve (AURC) for a given classification metric $\mathcal{A}$, defined as:
\begin{equation}
    \text{AURC-}\mathcal{A} = \int_0^1 \mathcal{A}(\mathcal{D}(r)) dr
\end{equation}
where $\mathcal{A}$ is any one of AUROC, Acc, or bAcc, and $\mathcal{D}(r)$ is the retained dataset consisting of samples with predictive entropy less than the $(1-r)$-quantile for standard referral, and given in equation (\ref{eqn:sr-retained}) for split referral. In practice, we approximate AURC by taking the mean of percentiles:
\begin{equation}
    \text{AURC-}\mathcal{A} = \frac{1}{R+1} \sum_{r=0}^{R} \mathcal{A} \left( \mathcal{D} \left( \frac{r}{100} \right) \right)
\end{equation}
where $R$ is the maximum referral rate. This was set at $R = 95$ because, for some referral methods, AUROC was undefined at very high referral rates (e.g., $r > 0.95$) when points of only one class remained.

Calibration performance is measured with expected calibration error (ECE).
We use the quantile binning strategy and set the number of bins, $B=15$.
\begin{equation}
    \text{ECE} = \frac{1}{n} \sum_{i=1}^{B} \left| \sum_{(x_j,y_j) \in \mathcal{B}_i} y_j - f_\theta(x_j) \right|
\end{equation}



\section*{C. Datasets and Preprocessing}
All datasets used in this paper are publicly available.
\begin{figure}[ht]
    \centering
    \begin{subfigure}{0.4\textwidth}
        \includegraphics[width=\textwidth]{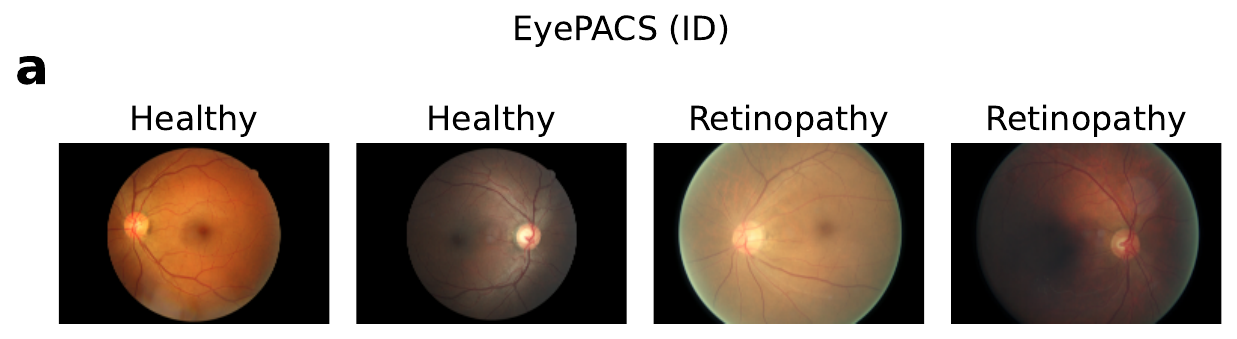}
    \end{subfigure}
    
    \begin{subfigure}{0.4\textwidth}
        \includegraphics[width=\textwidth]{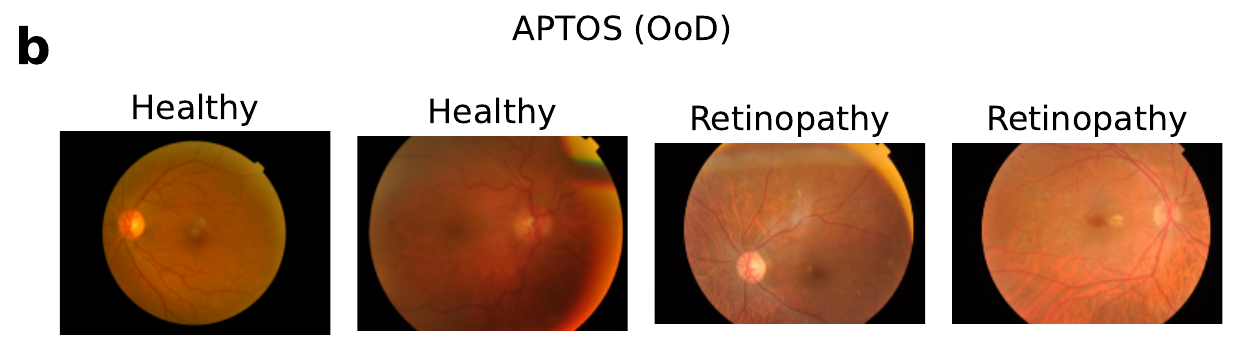}

    \end{subfigure}

    \caption{Diabetic Retinopathy: Exemplar images}

    \label{fig:dataset_retina}
\end{figure}

\begin{figure}[ht]
    \centering
    \begin{subfigure}{0.4\textwidth}
        \includegraphics[width=\textwidth]{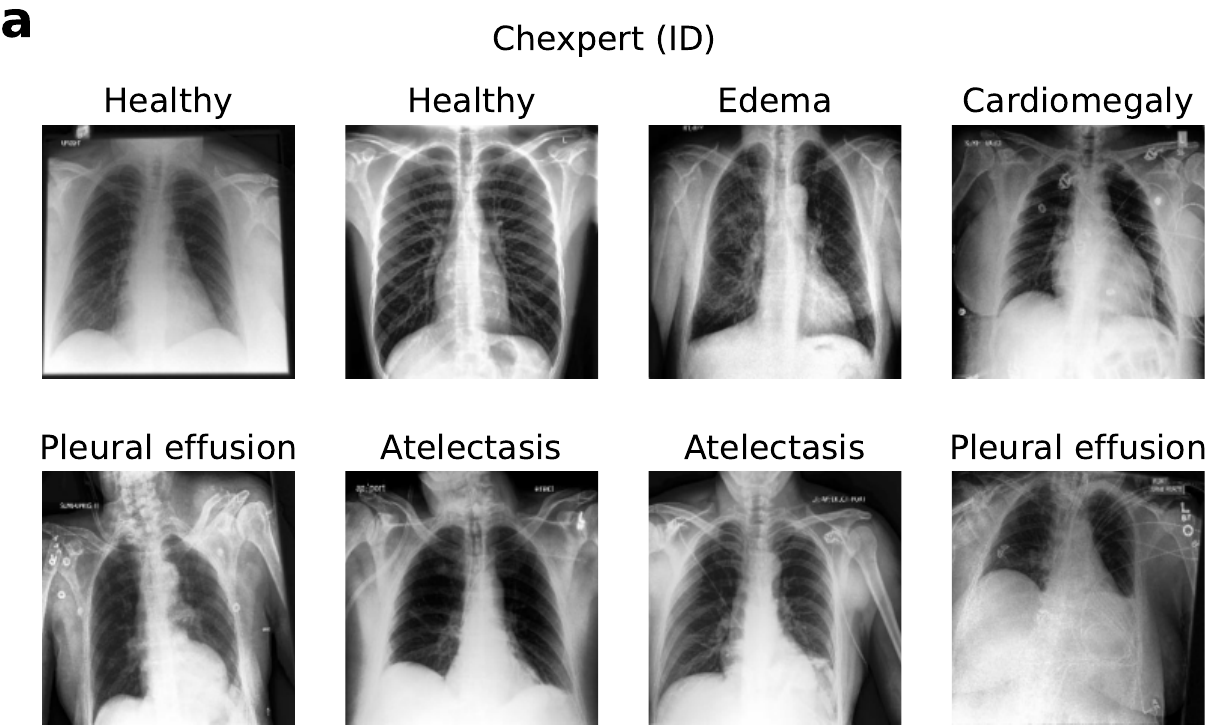}
    \end{subfigure}\\
    \begin{subfigure}{0.4\textwidth}
        \includegraphics[width=\textwidth]{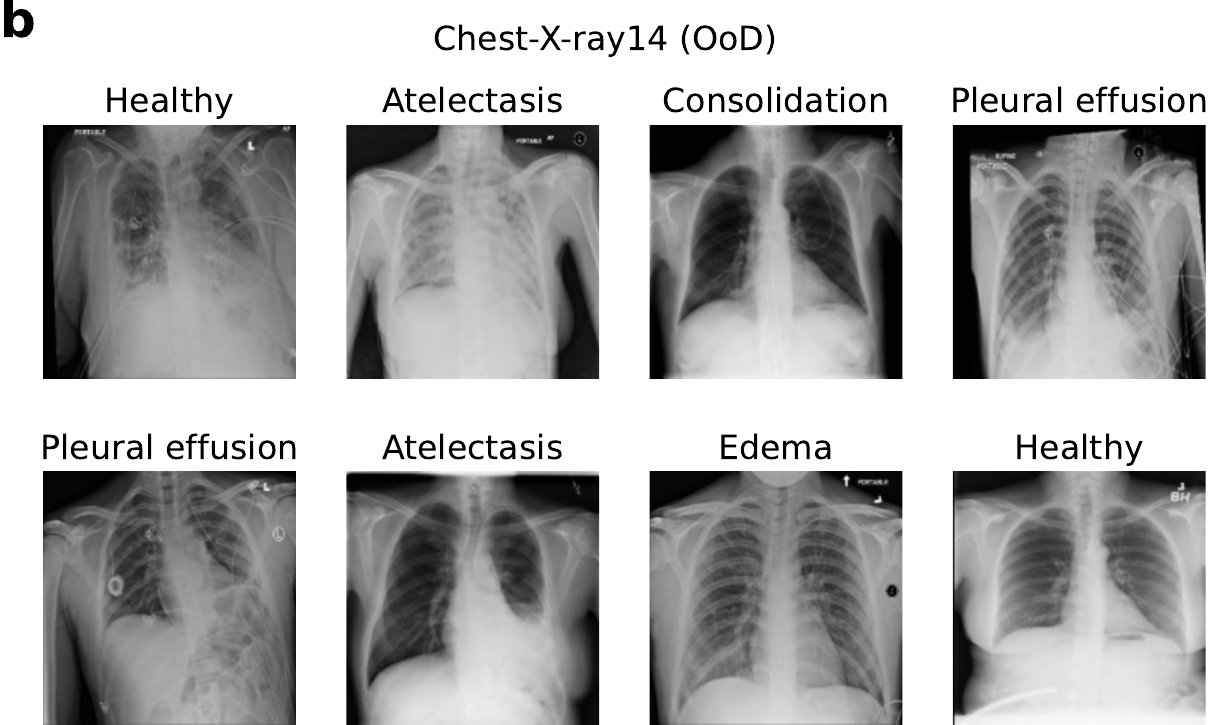}

    \end{subfigure}   

    \caption{Chest X-ray: Exemplar images}

    \label{fig:dataset_chest}
\end{figure}

\begin{table}[t]
\centering
\begin{tabular}{@{}ccr@{}}
\toprule
\textbf{Task}                & \textbf{\begin{tabular}[c]{@{}c@{}}Dataset (Domain)\end{tabular}} & \multicolumn{1}{c}{\textbf{Train / Val / Test}} \\ \midrule
\multirow{2}{*}{DR} & EyePacs (ID)                                                      & 35,126 / 10,906 / 42,670                        \\
                             & APTOS (OOD)                                                       & 733 / - / 2,929                                 \\ \midrule
\multirow{2}{*}{CX} & CheXpert (ID)                                                     & 191,027 / 202 / 518                             \\
                             & ChestX-ray14 (OOD)                                                & 89,795 / - / 22,325                             \\ \bottomrule 

\end{tabular}
\caption{DR and CX dataset details. Train split of the OOD dataset is used in self-supervised pretraining. OOD datasets do not have validation split. }
\label{tab:dataset}
\end{table}

\subsection*{C.1. Diabetic Retinopathy}
For studying referral with diabetic retinopathy (fundus) images, we employ the EyePACS dataset as the in-domain (ID) dataset, previously released as part of the Kaggle Diabetic Retinopathy Detection Challenge~\cite{eyepacs}. Each RGB image is labeled by a medical expert using the following labels: 0 (no diabetic retinopathy), 1
(mild diabetic retinopathy), 2 (moderate diabetic retinopathy), 3 (severe
diabetic retinopathy), and 4 (proliferative diabetic retinopathy). To binarize the labels, we divide these five labels into two groups (no/mild: labels {0,1} and moderate/severe/proliferative: labels {2,3,4}).
We employ the APTOS 2019 Blindness Detection dataset~\cite{aptos} for the out-of-domain (OOD) images  with the same labeling scheme as in EyePACS. The images of the APTOS dataset are noisier than those the EyePACS dataset, with distinct visual artifacts \cite{band2021benchmarking}. 

\noindent EyePACS can be downloaded with the Kaggle API: \\ \texttt{kaggle competitions download -c\\diabetic-retinopathy-detection} \\
APTOS can also be downloaded with the Kaggle API: \\ \texttt{kaggle competitions download -c\\aptos2019-blindness-detection} \\

For both datasets, we follow the pre-processing strategy employed by the winning entry of the Kaggle Diabetic Retinopathy Detection Challenge\footnote{\url{https://kaggle-forum-message-attachments.storage.googleapis.com/88655/2795/competitionreport.pdf}}. This includes resizing the image such that the retinal fundus  has a uniform radius of 300 pixels, Gaussian background subtraction (subtracting a Gaussian-filtered version of the image from the original), and clipping to 90\% size to avoid imaging artifacts at the image boundary. Finally, images are resized to $512\times512$ pixels. Batch size is set to 64. Figure-\ref{fig:dataset_retina} shows sample retinal fundus images from each dataset. 

\subsection*{C.2. Chest X-ray Conditions}
\label{subsec:appendix-cx-conditions}
For studying referral with chest X-ray images, we employ the CheXpert dataset ~\cite{irvin2019chexpert} as the ID dataset. This dataset is annotated based on 14 radiological findings (conditions). In CheXpert, labels of the train split were identified through NLP models from radiology reports, whereas labels of the validation and test set were manually annotated by board-certified radiologists; thus, labels of the train split are likely somewhat noisier than those of the test split. We examine 5 conditions previously analyzed in the original REMEDIS study~\cite{azizi2023}: atelectasis, consolidation, pulmonary edema, pleural effusion, and cardiomegaly. Training was performed for each label as binary classification (presence versus absence of that specific condition).
We employ the Chest X-ray14 dataset~\cite{Wang_2017_CVPR} as the OOD dataset, with the same five conditions as the ID, and labels extracted through NLP models from expert reports. 

CheXpert is publicly available for download\footnote{\url{https://stanfordaimi.azurewebsites.net/datasets/8cbd9ed4-2eb9-4565-affc-111cf4f7ebe2}}.
We filter out lateral chest X-ray scans, which comprise  $\sim$15\% of the dataset, following~\cite{azizi2023}.
ChestX-ray14 is also publicly available for download\footnote{\url{https://nihcc.app.box.com/v/ChestXray-NIHCC/folder/36938765345}}. A key difference with the REMEDIS study~\cite{azizi2023} is that we resample splits such that the proportion of train:test images is 4:1 in the OOD data, while ensuring that all scans from a single patient belong to only one of the two splits. We performed this resampling because we observed significant differences in classification and referral trends between the ID train and test data, indicative of substantially heterogeneous features within the OOD domain itself. 
Images are resized to $256\times256$ pixels, and batch size is set to 256. Figure-\ref{fig:dataset_chest} shows sample chest X-ray images from the ID and OOD datasets.

Data-loading scripts, annotation files, and formats for raw data are all included in our source codebase.


\section*{D. Split Referral Analysis}
In this section we analyze why and when split referral works. For this, we rewrite the model formulation as follows.

The random variable $z$ takes on the predicted probability of an input $x$ belonging to the positive class, given by the model $f_\theta(\cdot)$. The marginal density of $z$ from the data density $p_\mathcal{D}$ is given by:
\begin{equation}
p(z) = \int p_\mathcal{D}(x) p_\theta(z|x) dx
\end{equation}
where $p_\theta(z|x)$ is equal to the Dirac delta function $\delta(z - f_\theta(x))$, since the model is deterministic.

Its entropy value is given by:
\begin{equation}
\mathcal{H}(z) = - z \log z - (1 - z) \log (1 - z)
\end{equation}

We rank the samples separately for negative and positive predictions, by finding the empirical CDF over the entropy values:
\begin{equation}
\Phi_0(h) = \frac{\int_0^{0.5} p(z) \mathbb{I}[\mathcal{H}(z) < h] dz}{\int_0^{0.5} p(z) dz}
\label{eq:entropy-cdf-0}
\end{equation}

\begin{equation}
\Phi_1(h) = \frac{\int^1_{0.5} p(z) \mathbb{I}[\mathcal{H}(z) < h] dz}{\int^1_{0.5} p(z) dz}
\label{eq:entropy-cdf-1}
\end{equation}

The inverse entropy is defined as:
\begin{equation}
h = \mathcal{H}(z) \And 0 \le z \le 0.5 \implies z = \mathcal{H}^{-1}(h)
\end{equation}

For coverage rate $c$ (one minus referral rate), we determine thresholds over $z$ to specify which samples constitute the retained set $\mathcal{D}_\text{\tiny SR}(c)$, such that the CDF of the corresponding entropy is less than $c$.
\begin{equation}
z_0(c) = \mathcal{H}^{-1} ( \Phi_0^{-1}(c) );\ z_1(c) = 1 - \mathcal{H}^{-1} ( \Phi_1^{-1}(c) )
\end{equation}

\begin{equation}
\label{eqn:sr-retained}
\begin{split}
\mathcal{D}_\text{\tiny SR}(c) & = \left\{ x \in \mathcal{D} : \mathcal{H}(f_\theta(x)) < \Phi_{\mathbb{I}[f_\theta(x) \ge 0.5]}^{-1}(c) \right\} \\
                               & = \left\{ x \in \mathcal{D} : f_\theta(x) \in [0, z_0(c)] \cup [z_1(c), 1] \right\}
\end{split}
\end{equation}

In the main text, we describe sets of negative and positive predictions $\mathcal{N}$ and $\mathcal{P}$. The retained subsets of $\mathcal{N}$ and $\mathcal{P}$ contain $c|\mathcal{N}|$ and $c|\mathcal{P}|$ samples respectively due to the definitions of $z_0(c)$ and $z_1(c)$, thus maintaining the proportion of predicted labels.

The observed probability of a class conditioned on $z$ is as follows:
\begin{equation}
p(y|z) = \frac{1}{p(z)} \int p_\mathcal{D}(x) p(y|x) p_\theta(z|x) dx
\end{equation}

Note that in the calibration plot (Fig. 7d), the x-axis is $z$ and y-axis is $p(y=1|z)$, and $\text{ECE} = \mathbb{E}_{z \sim p(z)} \left[ |p(y=1|z) - z| \right]$. Precision values (for each class $y \in \{0, 1\}$) over the retained dataset are given by:
\begin{equation}
P_0(c) = \frac{\int_0^{z_0(c)} p(z, y=0) dz}{\int_0^{z_0(c)} p(z) dz}
\label{eq:precision-0}
\end{equation}

\begin{equation}
P_1(c) = \frac{\int^1_{z_1(c)} p(z, y=1) dz}{\int^1_{z_1(c)} p(z) dz}
\label{eq:precision-1}
\end{equation}

We show that the denominators in (\ref{eq:precision-0}) and (\ref{eq:precision-1}) are linear functions of $c$ using (\ref{eq:entropy-cdf-0}) and (\ref{eq:entropy-cdf-1}):
\begin{equation}
\begin{split}
\int_0^{z_0(c)} p(z) dz & = \int_0^{0.5} p(z) \mathbb{I}[z < z_0(c)] dz
\\                      & = \int_0^{0.5} p(z) \mathbb{I}[\mathcal{H}(z) < \Phi_0^{-1}(c)] dz
\\                      & = c \int_0^{0.5} p(z) dz = c K_0
\end{split}
\end{equation}

\begin{equation}
\begin{split}
\int^1_{z_1(c)} p(z) dz & = \int^1_{0.5} p(z) \mathbb{I}[z > z_1(c)] dz
\\                      & = \int^1_{0.5} p(z) \mathbb{I}[\mathcal{H}(z) < \Phi_1^{-1}(c)] dz
\\                      & = c \int^1_{0.5} p(z) dz = c K_1
\end{split}
\end{equation}

Next, we show that if the calibration plot is a monotonically increasing function, which is a property all models exhibit (Fig. \ref{fig:dan-eff}d), then both $P_0$ and $P_1$ are decreasing functions of $c$.
\begin{equation}
z_1 < z_2 \iff p(y=0|z_1) > p(y=0|z_2)
\label{eq:sr-bound-0}
\end{equation}
\begin{equation*}
\centering
\begin{split}
P_0(c) & = \frac{1}{c K_0} \int_0^{z_0(c)} p(z) p(y=0|z) dz
\\ & > \frac{1}{c K_0} \int_0^{z_0(c)} p(z) p(y=0|z_0(c)) dz
\\ & = p(y=0|z_0(c)) \frac{1}{c K_0} \int_0^{z_0(c)} p(z) dz
\\ & = p(y=0|z_0(c))
\end{split}
\end{equation*}

\begin{equation}
\therefore P_0(c) > p(y=0|z_0(c))
\label{eq:sr-bound-1}
\end{equation}

Now, let $c' > c$
\begin{equation*}
\begin{split}
K_0 c' P_0(c') & = \int_0^{z_0(c')} p(z) p(y=0|z) dz
\\
K_0 c' P_0(c') & = \int_0^{z_0(c)} p(z) p(y=0|z) dz
\\ & + \int_{z_0(c)}^{z_0(c')} p(z) p(y=0|z) dz
\end{split}
\end{equation*}
\begin{equation*}
\begin{split}
K_0 c' P_0(c') & = K_0 c P_0(c) + \int_{z_0(c)}^{z_0(c')} p(z) p(y=0|z) dz
\\
K_0 c' P_0(c') & = K_0 c' P_0(c) - K_0 (c' - c) P_0(c)
\\ & + \int_{z_0(c)}^{z_0(c')} p(z) p(y=0|z) dz
\end{split}
\end{equation*}
\begin{equation*}
\begin{split}
K_0 c' (P_0(c') - P_0(c)) & = - P_0(c) \int_{z_0(c)}^{z_0(c')} p(z) dz
\\ & + \int_{z_0(c)}^{z_0(c')} p(z) p(y=0|z) dz
\end{split}
\end{equation*}
\begin{equation*}
\begin{split}
& K_0 c' (P_0(c') - P_0(c))
\\ & = \int_{z_0(c)}^{z_0(c')} p(z) (p(y=0|z) - P_0(c)) dz
\\ & < \int_{z_0(c)}^{z_0(c')} p(z) (p(y=0|z) - p(y=0|z_0(c))) dz \ [\because (\ref{eq:sr-bound-1})]
\\ & < 0 \ [\because z_0(c) < z \text{ and } (\ref{eq:sr-bound-0})]
\end{split}
\end{equation*}

\begin{equation}
\therefore c' > c \implies P_0(c') < P_0(c)
\end{equation}
and similarly for $P_1$.

\begin{figure}[ht]
    \centering

    \includegraphics[width=0.47\textwidth]{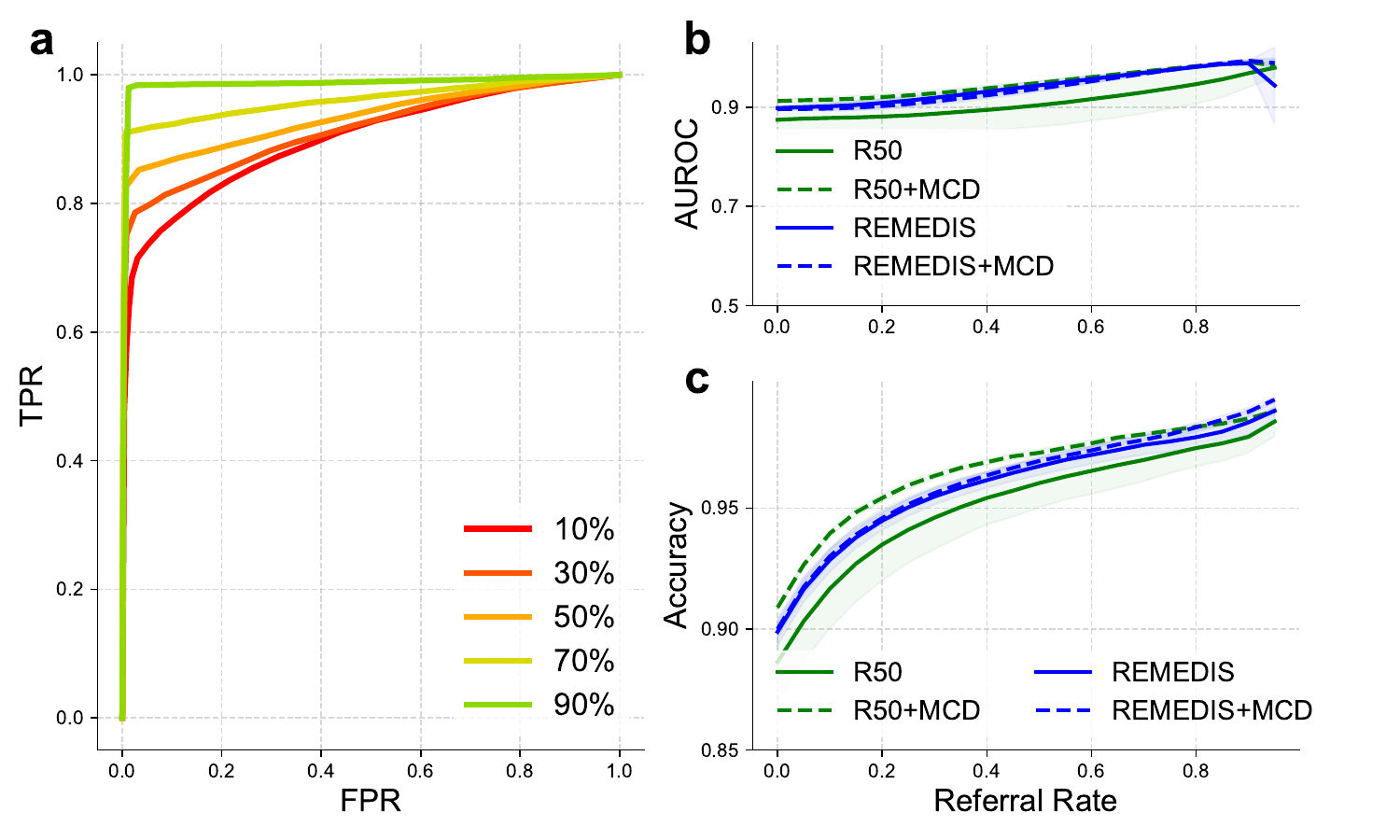}
    \caption{
        (a) ROCs for different referral rates (10-90\%) on the Diabetic Retinopathy ID (EyePACS) dataset with the baseline model (ResNet50).
        (b) and (c): Referral curves for AUROC (b) and Accuracy (c). See Fig. 2 for details.
    }
    \label{fig:rocs-resnet-id}
\end{figure}

\begin{table}[ht]
\centering
\begin{tabular}{lllll}
\toprule 
                                  & \multicolumn{2}{c}{\textbf{Retinopathy (ID)}}                                                                                                & \multicolumn{2}{c}{\textbf{Chest X-ray (ID)}}                                                                                                     \\
                                  
\multirow{-2}{*}{\textbf{Method}} & \textbf{\begin{tabular}[c]{@{}c@{}}AURC\\ AUROC\end{tabular}} & \textbf{\begin{tabular}[c]{@{}c@{}}AURC\\ Acc.\end{tabular}} & \textbf{\begin{tabular}[c]{@{}c@{}}AURC\\ AUROC\end{tabular}} & \textbf{\begin{tabular}[c]{@{}c@{}}AURC\\ B Acc.\end{tabular}} \\
\midrule
ViT-Base  &  0.9305  &  0.9585  &  0.9829  &  0.9617  \\
ViT+MCD   &  0.9362  &  \textbf{0.9623}$^*$  &  0.9869  &  0.9676  \\
\bl{ViT+OSP}   &  0.9389  &  0.9405  &  0.9826  &  0.9534  \\
\bl{ViT+SR}    &  0.9285  &  0.9565  &  0.9844  &  0.9650  \\
R/SimCLR  &  \textbf{0.9395}$^*$  &  0.9596  &  \textbf{0.9886}  &  \textbf{0.9676}$^*$  \\
\bl{R/SimMIM}  &  0.9286  &  0.9614  &  0.9828  &  0.9589  \\
\bl{R/MAE}     &  0.9223  &  0.9545  &  0.9802  &  0.9473  \\
\bl{R/DAN}     &  0.8998  &  0.9517  &  0.9857  &  0.9662  \\
\bl{ViT+IW}    &  0.9214  &  0.9552  &  0.9801  &  0.9564  \\
\bl{R/DAN+IW}  &  0.9214  &  0.9547  &  0.9781  &  0.9484  \\
\bottomrule
\end{tabular}
\caption{Model performance (AURC) on ID data:`R'- REMEDIS and `B Acc.' - Balanced Accuracy. 
\textbf{Bold} denotes the highest AURC. Scores are averaged over 6 seeds. $^*$difference between best and others is not significant (p$>$0.05).
}

\label{table:aurc-id}
\end{table}

\begin{table}[ht]
\centering
\begin{tabular}{lllll}
\toprule
\multirow{2}{*}{\textbf{Method}} & \multirow{2}{*}{\textbf{AUROC}} & \multirow{2}{*}{\textbf{Acc.}} & \textbf{Avg.} & \textbf{F1} \\
& & & \textbf{Prec.} & \textbf{score} \\
\midrule
ViT-Base  &  0.9513  &  0.8781  &  0.9038  &  0.8680  \\
ViT+MCD   &  \textbf{0.9609}  &  \textbf{0.8914}$^*$  &  \textbf{0.9273}  &  \textbf{0.8806}$^*$  \\
\bl{ViT+OSP}   &  0.9490  &  0.8512  &  0.9006  &  0.8448  \\
R/SimCLR  &  0.9470  &  0.8862  &  0.8943  &  0.8757  \\
\bl{R/SimMIM}  &  0.9564  &  0.8886  &  0.9198  &  0.8774  \\
\bl{R/MAE}     &  0.9505  &  0.8833  &  0.9001  &  0.8726  \\
\bl{R/DAN}     &  0.9490  &  0.8871  &  0.9036  &  0.8702  \\
\bl{ViT+IW}    &  0.9457  &  0.8852  &  0.8868  &  0.8740  \\
\bl{R/DAN+IW}  &  0.9475  &  0.8876  &  0.8935  &  0.8758  \\
\bottomrule                       
\end{tabular}
\caption{Diabetic Retinopathy OOD: various classification scores at zero referral rate. The outputs of ViT+SR are identical to ViT-Base at zero referral rates, hence all scores are also identical. $^*$difference between best and others is not significant (p$>$0.05).
}
\label{table:dr-ood-zero-ref}
\end{table}

\begin{table}[ht]
\centering
\begin{tabular}{lllll}
\toprule
\multirow{2}{*}{\textbf{Method}} & \multirow{2}{*}{\textbf{AUROC}} & \textbf{Bal.} & \textbf{Avg.} & \textbf{F1} \\
& & \textbf{Acc.} & \textbf{Prec.} & \textbf{score} \\
\midrule
ViT-Base  &  0.8226  &  0.7300  &  0.2683  &  0.3159  \\
ViT+MCD   &  0.8282  &  0.7314  &  \textbf{0.2782}  &  \textbf{0.3227}$^*$  \\
\bl{ViT+OSP}   &  0.8198  &  0.7236  &  0.2652  &  0.3142  \\
R/SimCLR  &  \textbf{0.8300}$^*$  &  0.7425  &  0.2756  &  0.3198  \\
\bl{R/SimMIM}  &  0.8235  &  0.7294  &  0.2595  &  0.3149  \\
\bl{R/MAE}     &  0.8054  &  0.7159  &  0.2429  &  0.2963  \\
\bl{R/DAN}     &  0.8235  &  \textbf{0.7502}$^*$  &  0.2683  &  0.3003  \\
\bl{ViT+IW}    &  0.8115  &  0.7299  &  0.2543  &  0.3003  \\
\bl{R/DAN+IW}  &  0.8104  &  0.7273  &  0.2536  &  0.2992  \\
\bottomrule                       
\end{tabular}
\caption{Chest X-ray OOD: various classification scores at zero referral rate. The outputs of ViT+SR are identical to ViT-Base, hence all scores are also identical. $^*$difference between best and others is not significant (p$>$0.05).
}
\label{table:cx-ood-zero-ref}
\end{table}

The total accuracy is a convex combination of the precisions for any coverage rate:

\begin{equation}
\begin{split}
\text{Acc}_\text{SR}(c) & = \frac{\int_0^{z_0(c)} p(z, y=0) dz + \int^1_{z_1(c)} p(z, y=1) dz}{\int_0^{z_0(c)} p(z) dz + \int^1_{z_1(c)} p(z) dz}
\\                    & = \frac{c K_0 P_0(c) + c K_1 P_1(c)}{c K_0 + c K_1}
\\                    & = \frac{K_0 P_0(c) + K_1 P_1(c)}{K_0 + K_1}
\end{split}
\end{equation}

\begin{table*}[t]
\centering
\begin{tabular}{lllllllllll}
\toprule
\multirow{3}{*}{\textbf{Method}} & \multicolumn{2}{c}{\textbf{Atelectasis}}                                                                                                           & \multicolumn{2}{c}{\textbf{Consolidation}}                                                                                                         & \multicolumn{2}{c}{\textbf{Edema}}                                                                                                                 & \multicolumn{2}{c}{\textbf{Effusion}}                                                                                                              & \multicolumn{2}{c}{\textbf{Cardiomegaly}}                                                                                                          \\
                                 & \textbf{\begin{tabular}[c]{@{}c@{}}AURC\\ AUROC\end{tabular}} & \multicolumn{1}{c}{\textbf{\begin{tabular}[c]{@{}c@{}}AURC\\ B. Acc\end{tabular}}} & \textbf{\begin{tabular}[c]{@{}c@{}}AURC\\ AUROC\end{tabular}} & \multicolumn{1}{c}{\textbf{\begin{tabular}[c]{@{}c@{}}AURC\\ B. Acc\end{tabular}}} & \textbf{\begin{tabular}[c]{@{}c@{}}AURC\\ AUROC\end{tabular}} & \multicolumn{1}{c}{\textbf{\begin{tabular}[c]{@{}c@{}}AURC\\ B. Acc\end{tabular}}} & \textbf{\begin{tabular}[c]{@{}c@{}}AURC\\ AUROC\end{tabular}} & \multicolumn{1}{c}{\textbf{\begin{tabular}[c]{@{}c@{}}AURC\\ B. Acc\end{tabular}}} & \textbf{\begin{tabular}[c]{@{}c@{}}AURC\\ AUROC\end{tabular}} & \multicolumn{1}{c}{\textbf{\begin{tabular}[c]{@{}c@{}}AURC\\ B. Acc\end{tabular}}} 
\\
\midrule
ViT-Base  &  0.7763  &  \textbf{0.7297}$^*$
          &  0.7500  &  0.6595
          &  0.9086  &  0.7437
          &  0.8771  &  0.8338
          &  0.8694  &  0.8018 \\

ViT+MCD   &  0.7684  &  0.7066
          &  0.7022  &  0.6130
          &  0.9175  &  0.7513
          &  0.8474  &  0.7839
          &  \textbf{0.9004}$^*$  &  0.8224 \\

\bl{ViT+OSP}   &  0.7339  &  0.6707
          &  0.6913  &  0.6228
          &  0.8848  &  0.7174
          &  0.8740  &  0.8363
          &  0.8370  &  0.7017 \\

\bl{ViT+SR}    &  0.8025  &  0.7647
          &  \textbf{0.8643}$^\text{\#}$  &  \textbf{0.8311}$^\text{\#}$
          &  0.9340  &  0.9114
          &  \textbf{0.9180}$^\text{\#}$  &  0.8886
          &  0.8806  &  0.8256 \\

R/SimCLR  &  \textbf{0.8159}$^*$  &  0.7644
          &  0.7322  &  0.6677
          &  0.8534  &  0.7458
          &  0.8524  &  0.8148
          &  0.8711  &  0.8087 \\

R/SimMIM  &  0.7892  &  0.7373
          &  0.7336  &  0.6482
          &  0.9122  &  0.7411
          &  0.8292  &  0.7815
          &  0.8480  &  0.7697 \\

\bl{R/MAE}     &  0.7177  &  0.6572
          &  0.6982  &  0.6034
          &  0.8967  &  0.7418
          &  0.7725  &  0.7166
          &  0.8169  &  0.7506 \\

\bl{R/DAN}     &  0.8033  &  0.7670
          &  0.8437  &  0.8049
          &  \textbf{0.9567}  &  \textbf{0.9311}
          &  0.9162  &  \textbf{0.8890}$^\text{\#}$
          &  0.8885  &  \textbf{0.8397}$^*$ \\

\bl{ViT+IW}    &  0.7717  &  0.7214
          &  0.7201  &  0.6593
          &  0.9027  &  0.8177
          &  0.8580  &  0.8234
          &  0.8592  &  0.7851 \\

\bl{R/DAN+IW}  &  0.7763  &  0.7272
          &  0.7266  &  0.6740
          &  0.8829  &  0.7518
          &  0.8577  &  0.8210
          &  0.8430  &  0.7565 \\

\bottomrule
\end{tabular}
\caption{
Condition-wise AURC-AUROC and AURC-Balanced Accuracy on Chest X-Ray OOD (Chest-X-ray14) data. For other details, see Table 1 in the main text. $^*$difference between best and others is not significant (p$>$0.05). $^\text{\#}$difference between REMEDIS/DAN and ViT+SR is not significant (p$>$0.05). Thus, REMEDIS/DAN and ViT+SR outperform others on consolidation, edema, and effusion. On atelectasis and cardiomegaly, no method is clearly best.
}
\label{table:aurc-cx-cond-ood}
\end{table*}

Because both $P_0$ and $P_1$ are decreasing functions of $c$, their convex combination Acc$_\text{SR}$ is also a decreasing function of $c$ and, equivalently, an increasing function of referral rate. Thus SR is guaranteed to improve models which suffer referral failure due to miscalibration but whose calibration plots are monotonically increasing (e.g., Fig. \ref{fig:dan-eff}d). Future work will address SR failure cases: conditions  under which the calibration plot may not increase monotonically (e.g., depending on the severity of covariate shift).



\section*{E. Additional Experiments}

\subsection*{E.1. In-domain referral and OOD zero-referral classification performance}
We tested the baseline model (Resnet50) on the Diabetic Retinopathy ID (EyePACS) dataset by plotting ROCs for different referral rates (10-90\%) (Fig.~\ref{fig:rocs-resnet-id}a). Unlike with the OOD data, ROC increases as we increase the referral rate. In addition, referral curves of AUROC and Accuracy also increase monotonically  (Fig.~\ref{fig:rocs-resnet-id}b-c). Table~\ref{table:aurc-id} confirms that none of the evaluated approaches suffer from referral failures on the ID data (both CX and DR).

We also report various classification scores at zero referral, namely AUROC, (balanced) accuracy, average precision (equivalent to area under precision recall curve), and F1-score: in Table \ref{table:dr-ood-zero-ref} for DR and Table \ref{table:cx-ood-zero-ref} for CX. These vary marginally across the different models.

\subsection*{E.2. Simulation description}
We describe the synthetic data mentioned in Section \ref{sec:motivn}. Both ID and OOD samples are distributed as bivariate Gaussian mixtures. For both data domains, feature 1 has $\mu=0$ for class A and $\mu=5$ for class B. For both classes, feature 2 has $\mu=0$ for ID data and $\mu=5$ for OOD data. 

In each of the four mixture components, the covariance between features 1 and 2 is $\Sigma_{12}=0.4$, and their variances are $\Sigma_{11}=\Sigma_{22}=1$. In this sense, one feature is relevant for distinguishing classes (feature 1), and an orthogonal feature is relevant for distinguishing domains (feature 2). 

The classification boundary in Fig. \ref{fig:toy-lr}a is obtained by fitting logistic regression using scikit-learn\footnote{\url{https://scikit-learn.org/stable/modules/generated/sklearn.linear_model.LogisticRegression}}, with 1000 points sampled for each class in each domain. Because of the covariance among the features, the classifier boundary based on ID data alone does not optimally generalize to OOD data, yielding poor accuracy as well as miscalibrated predictions.


\subsection*{E.3. Measuring domain segregation}
We show that, following SimCLR contrastive pretraining (i.e, after REMEDIS step ii), ID and OOD DR data distributions get even more segregated than before such pretraining (i.e, after step i). We quantify this segregation with two different approaches. 

First, we computed the pairwise Euclidean distances of ViT's penultimate layer embeddings after PCA dimensionality reduction (2 components). We then computed the ratio of average inter-domain distances (subsample of 250 points each, across ID and OOD classes) to average intra-domain distances (subsamples of 250 points each, within the respective ID or OOD class); this procedure was repeated 10 times with different pseudorandom seeds, and resulting ratios were averaged. The resultant ratio was 1.06 after step i (pre-SimCLR), and significantly higher, at 1.19 after step ii (post-SimCLR) (p $< 10^{-12}$, two-sided \textit{t}-test). In other words, after SimCLR pretraining inter-domain distances increased significantly relative to intra-domain distances.

Second, we trained a logistic regression model to predict domain labels following each step, using the first two principal component features. Domain prediction accuracy was 63\% with REMEDIS step i (pre-SimCLR) representations and significantly higher, at 74\%, with step ii (post-SimCLR) representations (p $< 10^{-8}$, two-sided \textit{t}-test; 2-fold cross-validation, split sampled randomly 10 times). In other words, the ID and OOD data distributions became more separable after SimCLR.

Both analyses provide converging evidence for significant additional segregation of ID samples further away from OOD samples, following SimCLR.

\begin{figure*}[ht]
    \centering
    \begin{subfigure}{\textwidth}
        \includegraphics[width=\textwidth]{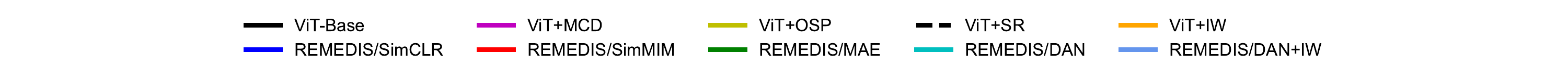}
    \end{subfigure}
    \\
    \begin{subfigure}{0.19\textwidth}
        \includegraphics[width=\textwidth]{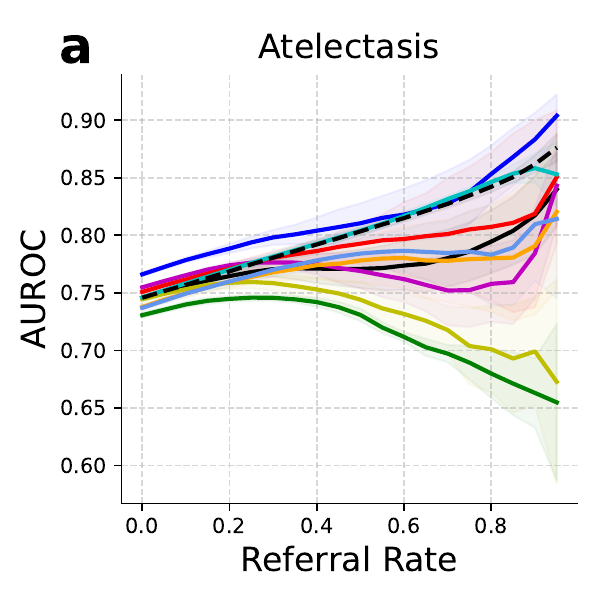}
    \end{subfigure}
    \begin{subfigure}{0.19\textwidth}
        \includegraphics[width=\textwidth]{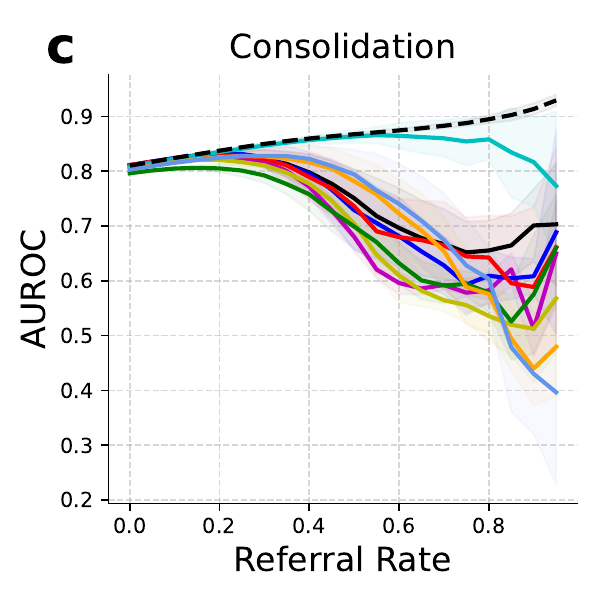}
    \end{subfigure}
    \begin{subfigure}{0.19\textwidth}
        \includegraphics[width=\textwidth]{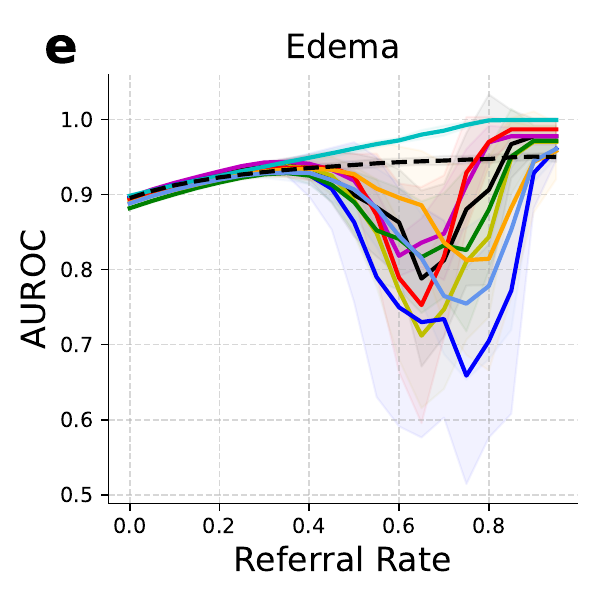}
    \end{subfigure}
    \begin{subfigure}{0.19\textwidth}
        \includegraphics[width=\textwidth]{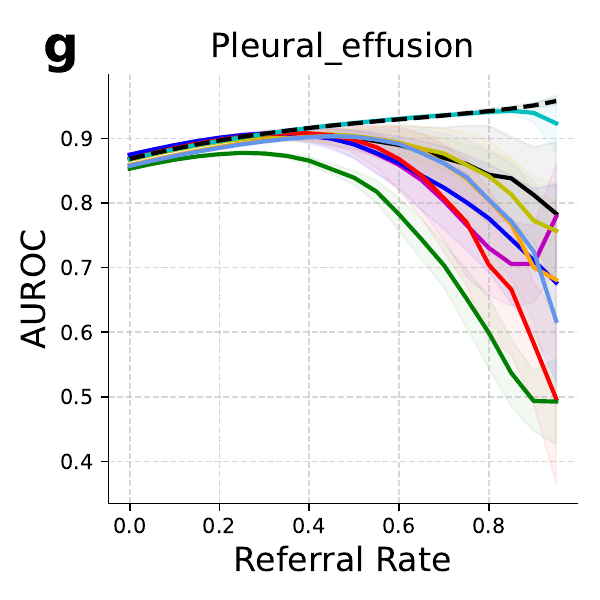}
    \end{subfigure}
    \begin{subfigure}{0.19\textwidth}
        \includegraphics[width=\textwidth]{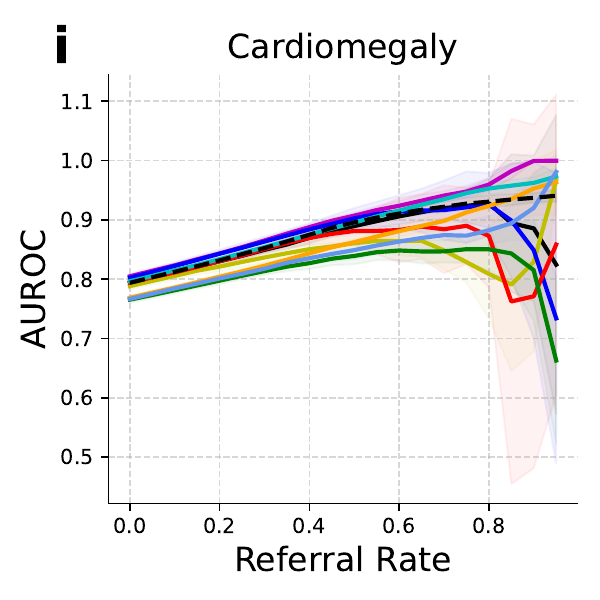}
    \end{subfigure}
    
    \begin{subfigure}{0.19\textwidth}
        \includegraphics[width=\textwidth]{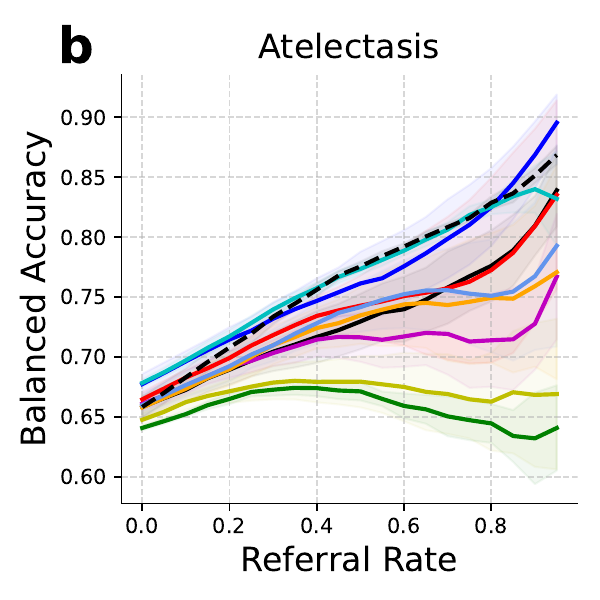}
    \end{subfigure}
    \begin{subfigure}{0.19\textwidth}
        \includegraphics[width=\textwidth]{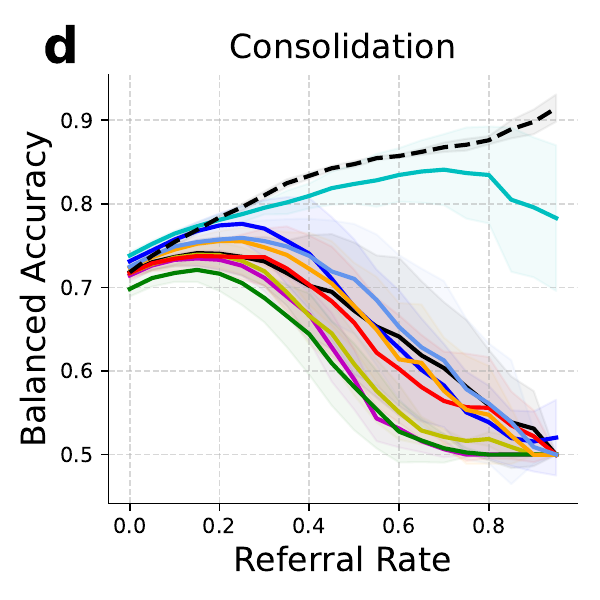}
    \end{subfigure}
    \begin{subfigure}{0.19\textwidth}
        \includegraphics[width=\textwidth]{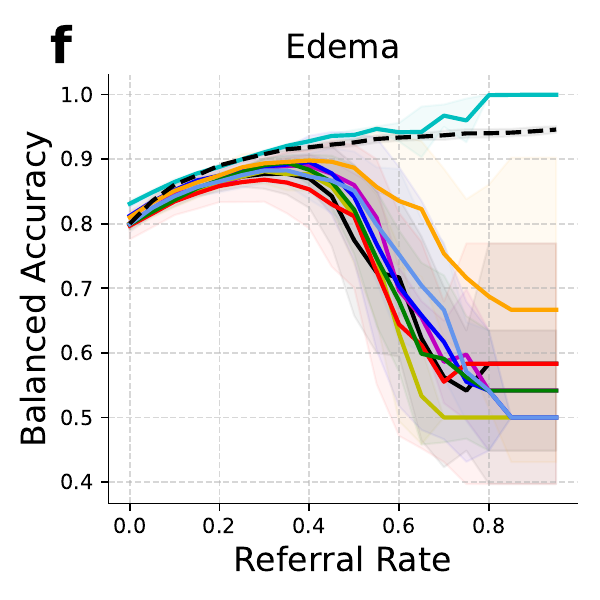}
    \end{subfigure}
    \begin{subfigure}{0.19\textwidth}
        \includegraphics[width=\textwidth]{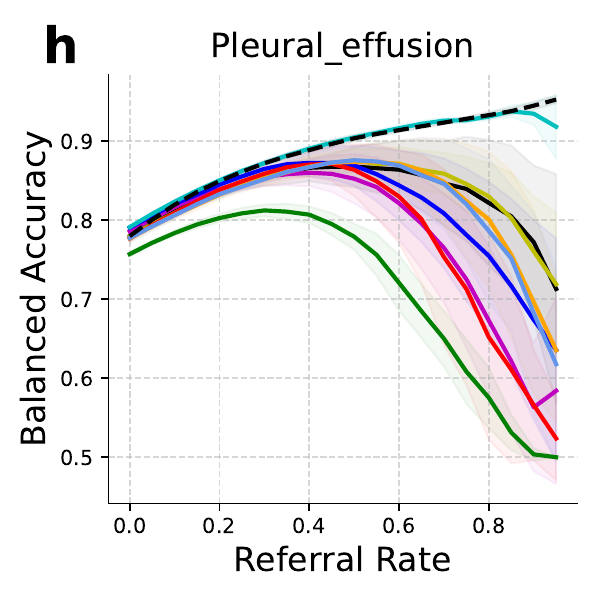}
    \end{subfigure}
    \begin{subfigure}{0.19\textwidth}
        \includegraphics[width=\textwidth]{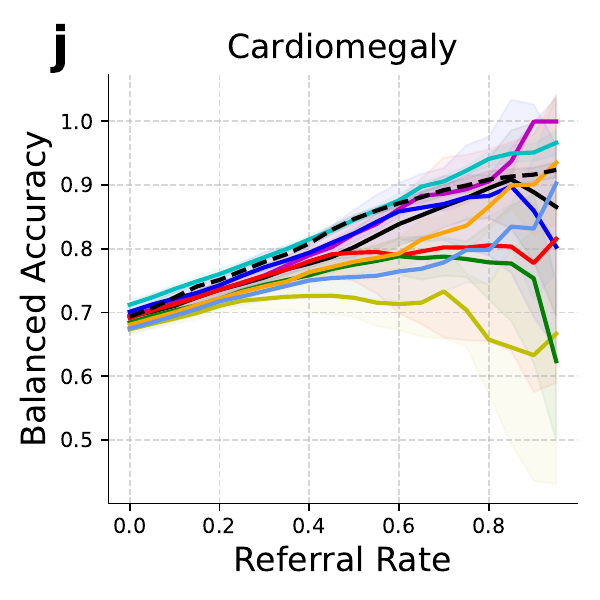}
    \end{subfigure}

    \caption{
        Referral performance on OOD-Chest X-ray Conditions: (top row: AUROC, bottom row: Balanced Accuracy) for each of the 5 conditions (columns). 
    }
    \label{fig:cx-conds-ref}
\end{figure*}

\begin{table*}[t]
\centering
\begin{tabular}{lccccc}
\toprule
\textbf{Method} & \textbf{Atelectasis} & \textbf{Consolidation} & \textbf{Edema} & \textbf{Effusion} & \textbf{Cardiomegaly} \\
\midrule
ViT-Base        &  0.2129  &  0.3082  &  0.1907  &  0.1172  &  0.1976 \\
ViT+MCD         &  0.2161  &  0.3135  &  0.1864  &  0.1169  &  0.2063 \\
REMEDIS/SimCLR  &  0.2026  &  0.2807  &  0.1599  &  0.1202  &  0.1964 \\
\bl{REMEDIS/SimMIM}  &  0.2136  &  0.3148  &  0.2062  &  0.1287  &  0.2019 \\
\bl{REMEDIS/MAE}     &  0.2305  &  0.3222  &  0.1956  &  0.1534  &  0.1940 \\
\bl{REMEDIS/DAN}     &  \textbf{0.1513}  &  \textbf{0.2635}  &  \textbf{0.1322}  &  \textbf{0.0736}  &  \textbf{0.1415} \\
\bl{ViT+IW}          &  0.1973  &  0.3053  &  0.1976  &  0.1074  &  0.1951 \\
\bl{REMEDIS/DAN+IW}  &  0.1813  &  0.2938  &  0.1932  &  0.0926  &  0.1944 \\
\bottomrule                       
\end{tabular}
\caption{Expected Calibration Error with 15 bins. REMEDIS/DAN achieves lower ECE than other methods (p$<$0.05, two-sided \textit{t}-test)}
\label{table:ECE-cx-ood}
\end{table*}

\begin{figure*}[t]
    \centering
    \includegraphics[width=\textwidth]{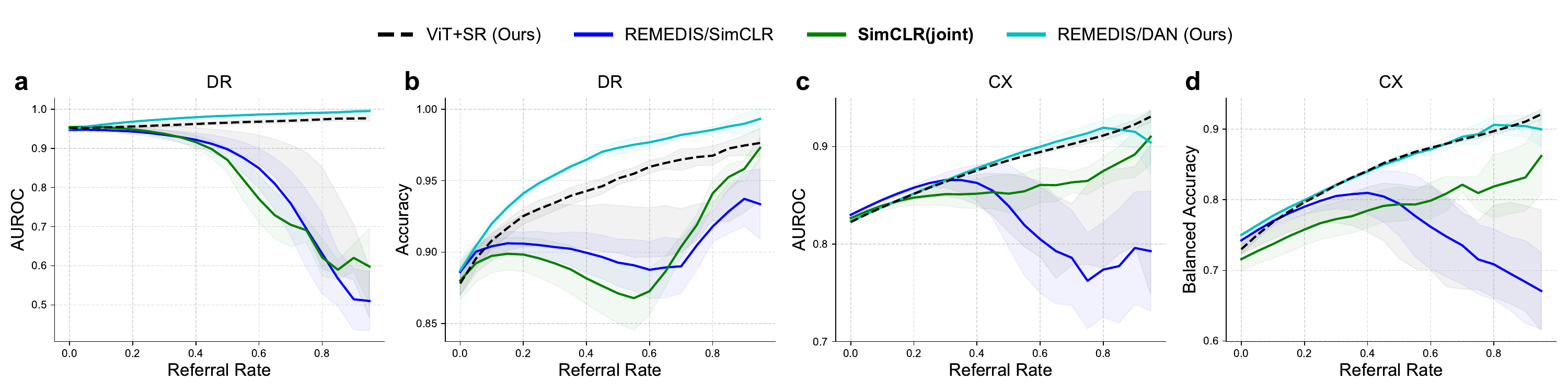}
    \caption{SimCLR(joint) referral curves on DR (a-b) and CX (c-d) data based on AUROC (a, c) and Accuracy (b, d). Other conventions are the same as in Fig. 6.}
    \label{fig:ref-simclr-joint}
\end{figure*}

\subsection*{E.4. CX Results: Condition-wise}

In section~\ref{subsec:appendix-cx-conditions}, we introduced five conditions of the Chest X-ray datasets on which we perform multilabel binary classification. Performance metrics reported in the main text (section 4.3) were averaged across the 5 conditions. To assess how the approaches perform on each condition, we further evaluate the binary classification model for each condition (a total of five models). For this evaluation, for each condition ``c'', we assign a positive class if disease ``c'' is present and assign a negative class if none of the 5 diseases are present. 

Overall, non-monotonicity of the referral curves is evident for 3 (``Consolidation'', ``Edema'', ``Effusion'') of the 5 conditions (Fig.~\ref{fig:cx-conds-ref}). For individual conditions, we observe improvements up to 26\% and 15\% in AURC-Balanced Accuracy and AURC-AUROC, respectively, over ViT-Base (Table~\ref{table:aurc-cx-cond-ood}). 

Next, we examine the most successful and least successful cases based on their relative improvements over the baseline. In the most-successful case -- Consolidation -- all the methods (except ours) exhibit markedly decreasing AUROC and accuracy referral curves after a $\sim$0.25 referral rate, yielding low AURC-AUROCs (0.69-0.75) (Table~\ref{table:aurc-cx-cond-ood}, columns 4-5). On the other hand, ViT+SR and REMEDIS/DAN achieved an AURC-AUROC of 0.86 and 0.84 with a relative improvement of 15.2\% and 12.5\%, respectively, over the baseline ViT model (Table~\ref{table:aurc-cx-cond-ood}, rows 4 \& 7). In addition, REMEDIS/DAN and ViT+SR achieved improvements ranging from 15-25\% over all other competing approaches. 

In two conditions -- ``Atelectasis'' and ``Cardiomegaly'' -- one of the competing methods outperformed either ViT+SR or REMEDIS/DAN. We observed that, in these two cases, only a few methods showed marked referral failures (Fig~\ref{fig:cx-conds-ref}a and e), yielding only marginal differences with REMEDIS/DAN and ViT+SR in terms of AURC.
In the Atelectasis case, REMEDIS/SimCLR achieved the highest AURC-AUROC, albeit by a small margin (relative improvement of 1.6\% over both ViT+SR \& REMEDIS/DAN) (Table~\ref{table:aurc-cx-cond-ood}, column 2).
Similarly, in the Cardiomegaly case, ViT+MCD outperformed ViT+SR \& REMEDIS/DAN in terms of AURC-AUROC, but again by small margins 2.2\% and 1.3\%, respectively) (Table~\ref{table:aurc-cx-cond-ood}, column 10).

As discussed in Section 4.4, REMEDIS/DAN significantly improved calibration on OOD data, which yielded better referral AURCs. Across the different CX conditions, REMEDIS/DAN consistently reduced ECE over ViT-Base, even for the diseases where we did not observe major differences in AURC. Among the 5 conditions, REMEDIS/DAN achieved varying reductions in ECE, ranging from 14\% for Consolidation (REMEDIS/DAN's ECE is 0.26, ViT's is 0.31) to 39\% for Pleural Effusion (REMEDIS/DAN's ECE is 0.07, ViT's is 0.12) (Table~\ref{table:ECE-cx-ood}).

\begin{figure*}[t]
    \centering
    \includegraphics[width=0.8\textwidth]{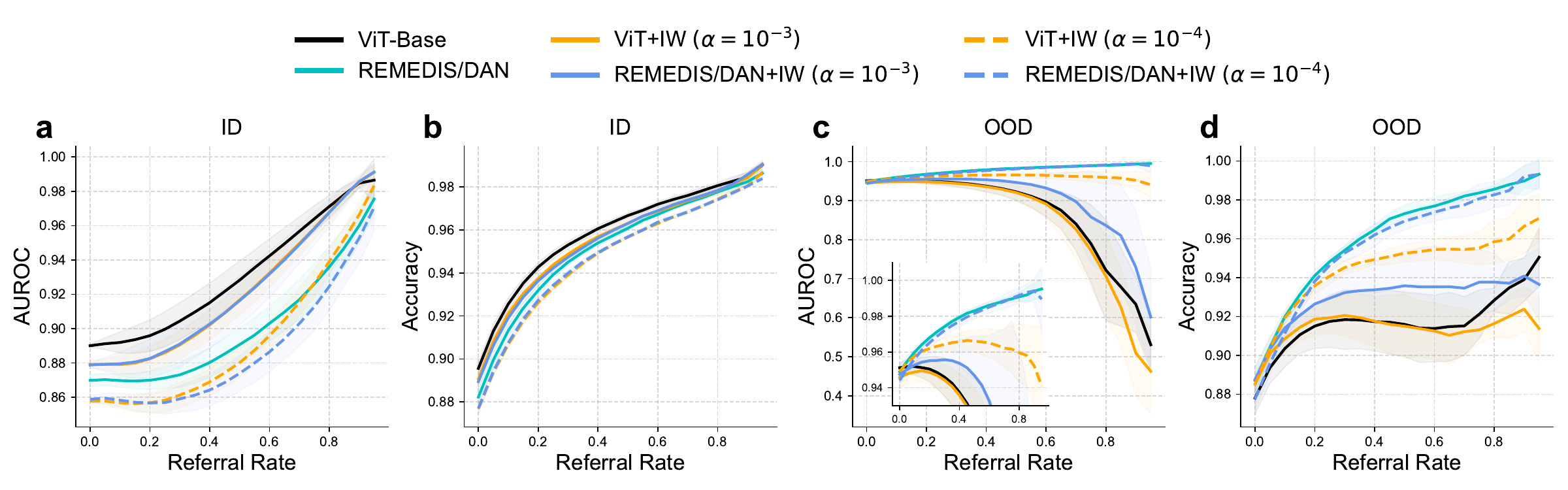}
    \caption{
        Referral curves for the DR datasets showing referral performance of the Importance Weighting models (ViT+IW and REMEDIS/DAN+IW) for ID data (a-b) and OOD data (c-d). 
        Solid orange and blue curves denote performance for the ViT+IW and REMEDIS/DAN+IW models, respectively, with the default learning rates $(\alpha = 10^{-3})$. Dashed orange and blue curves reflect performance for the same two models but with  a lower value of learning rate $(\alpha = 10^{-4})$.
    }
    \label{fig:iw-llr}
\end{figure*}

\subsection*{E.5. SimCLR(joint) Model}
In the REMEDIS framework, fine-tuning on the ID data (step iii) after the REMEDIS/SimCLR step (step ii) can override the domain generalization effect of the latter step. Thus, we tried jointly optimizing steps ii and iii. This ``SimCLR(joint)'' model has the same network architecture as REMEDIS/SimCLR, but both classification cross-entropy and the SimCLR loss are minimized jointly during training. In Fig.~\ref{fig:ref-simclr-joint}, we compared SimCLR(joint)'s referral curves on OOD datasets (DR and CX) with the default sequential model (REMEDIS/SimCLR) as well as ViT+SR and REMEDIS/DAN. In the case of DR, we observed that SimCLR(joint) also suffers from referral failure, much like REMEDIS/SimCLR (Fig.~\ref{fig:ref-simclr-joint}a-b). But, interestingly, in the case of CX referral failures are largely ameliorated Fig.~\ref{fig:ref-simclr-joint}c-d. Nevertheless, ViT+SR and REMEDIS/DAN perform better than SimCLR(joint) for both datasets. 


\subsection*{E.6. Entropy and logit distributions}
\begin{figure*}[ht]
    \centering
\begin{subfigure}{\textwidth}
    \includegraphics[width=\textwidth]{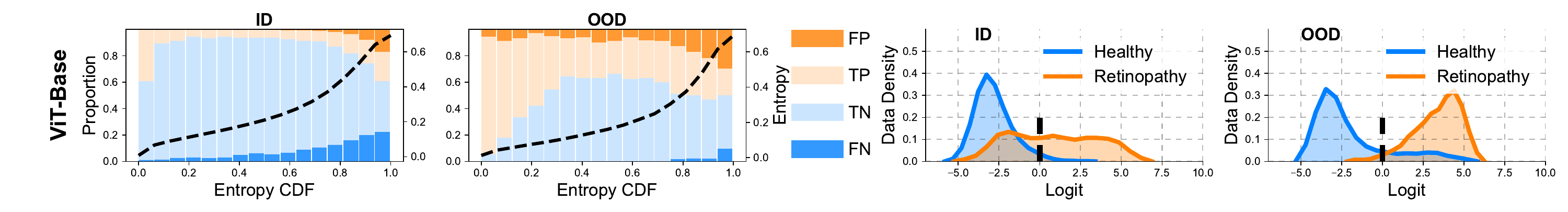}
\end{subfigure}

\begin{subfigure}{\textwidth}
    \includegraphics[width=\textwidth]{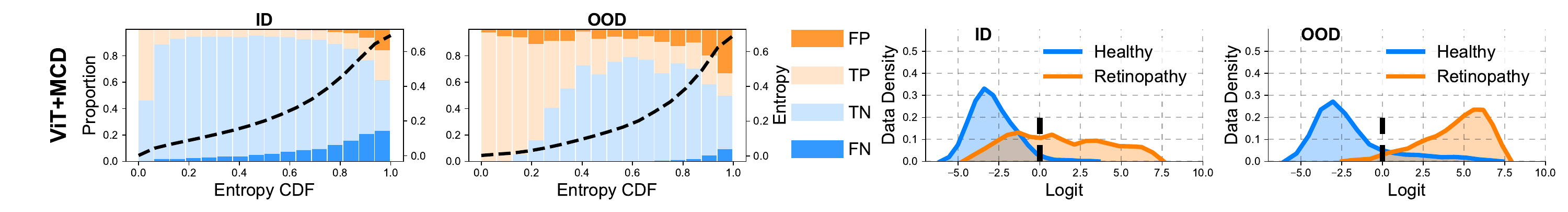}
\end{subfigure}

\begin{subfigure}{\textwidth}
    \includegraphics[width=\textwidth]{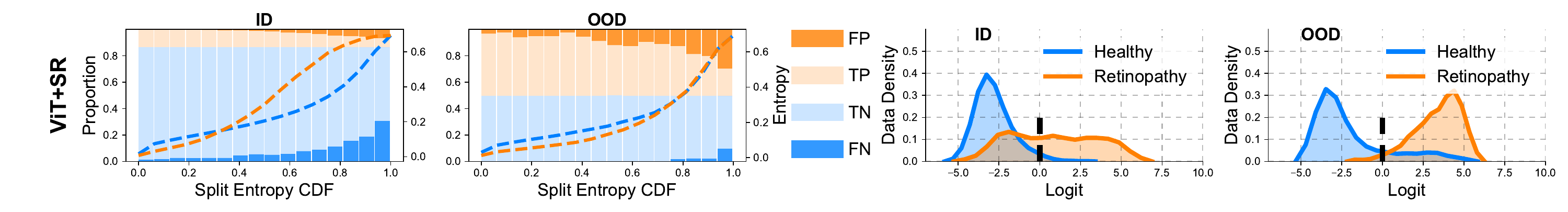}
\end{subfigure}

\begin{subfigure}{\textwidth}
    \includegraphics[width=\textwidth]{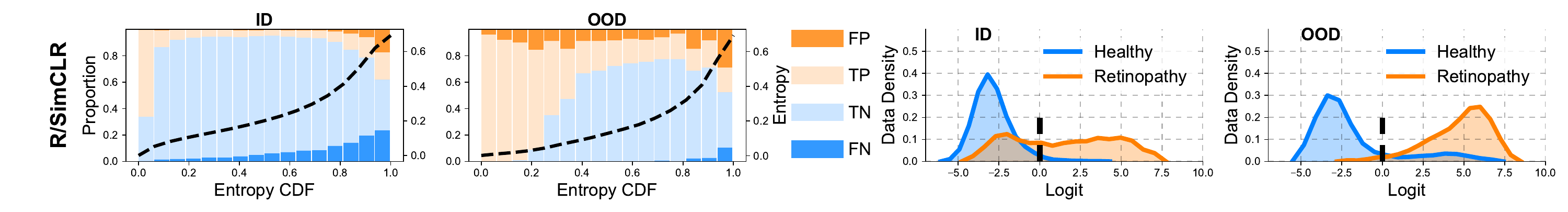}
\end{subfigure}

\begin{subfigure}{\textwidth}
    \includegraphics[width=\textwidth]{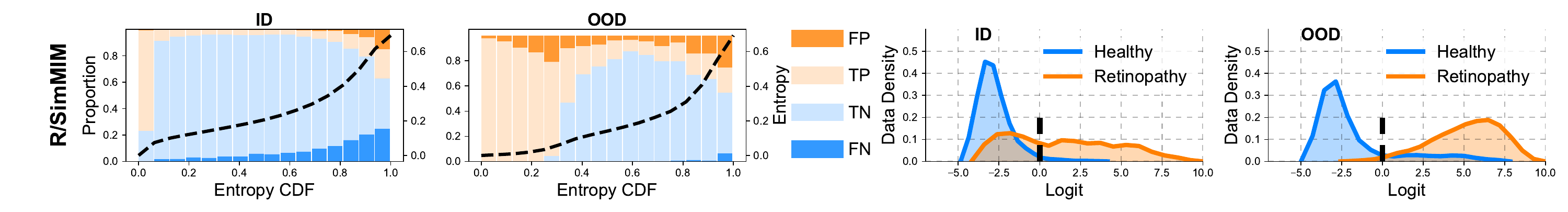}
\end{subfigure}

\begin{subfigure}{\textwidth}
    \includegraphics[width=\textwidth]{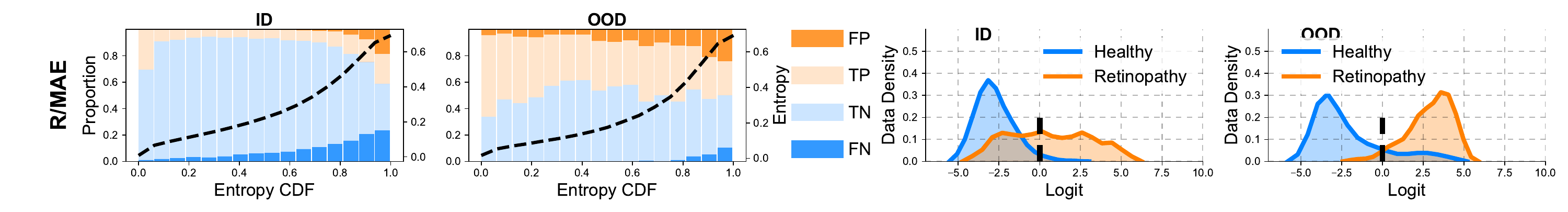}
\end{subfigure}

\begin{subfigure}{\textwidth}
    \includegraphics[width=\textwidth]{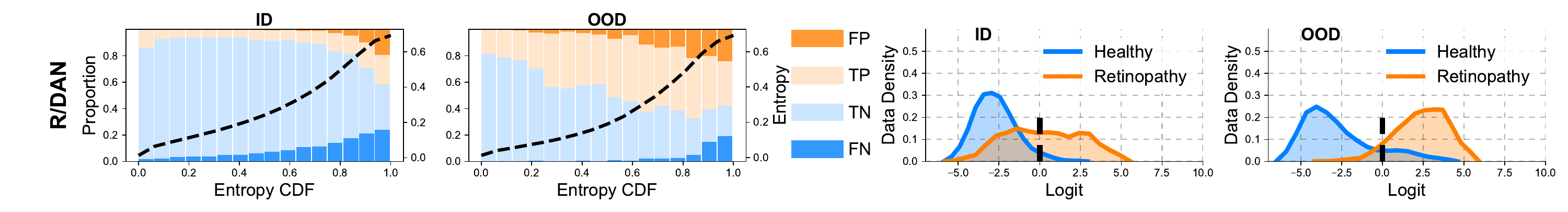}
\end{subfigure}

\begin{subfigure}{\textwidth}
    \includegraphics[width=\textwidth]{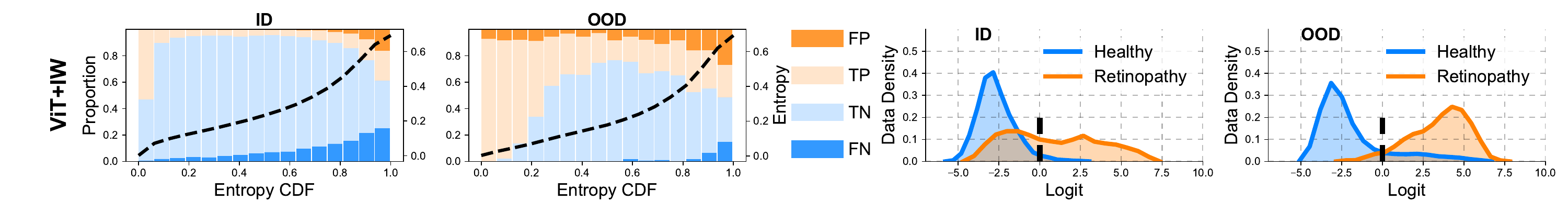}
\end{subfigure}

\begin{subfigure}{\textwidth}
    \includegraphics[width=\textwidth]{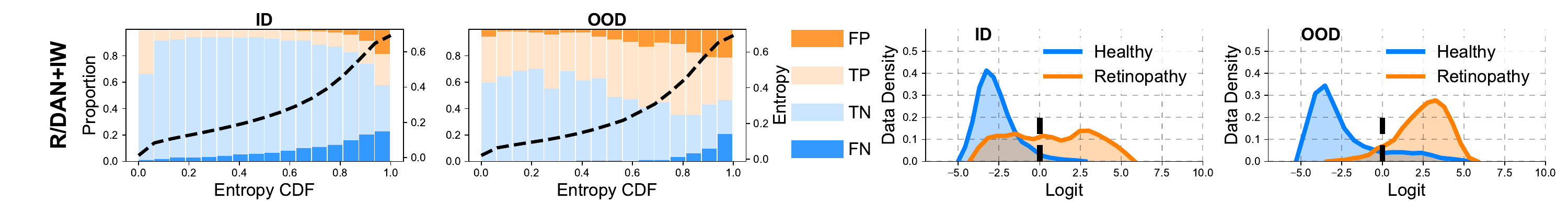}
\end{subfigure}

\caption{Entropy and Logit distributions for the different models for the DR data. Each row represents a particular model (label, leftmost). The columns represent entropy histograms for the ID (first column) and OOD (second column) data, as well as the logit histograms for the ID (third column) and OOD (fourth column) data. For other conventions, see Fig. 4. }
\label{fig:logits-entropy-all}
\end{figure*}

We visualize the predictive entropy and logit distributions of the different models -- ViT-Base, ViT+MCD,  ViT+SR, REMEDIS/SimCLR, REMEDIS/SimMIM, REMEDIS/MAE, REMEDIS/DAN, ViT+IW, and REMEDIS/DAN+IW -- to understand the reasons for their differential performance (Fig. \ref{fig:logits-entropy-all}). ViT+OSP gives separate probabilities for each class and is not directly comparable to the other models; it is, therefore, not included in the analysis. 

High entropy (right side of the plot) denotes high predictive uncertainty. Therefore, errors (FN and FP, Fig. \ref{fig:logits-entropy-all}, columns 1-2, dark bars) should be at more at the higher entropy regions. This pattern which is evident in the ID entropy distribution for all the models (Fig. \ref{fig:logits-entropy-all}, first column). But for OOD, we observe FPs in regions of lower entropy, i.e. toward the left side of the plot (e.g, Fig. \ref{fig:logits-entropy-all} first row, second column). This is due to the long tail of OOD logits with actual healthy labels, but which are predicted as diseased (long tail of the blue distributions in Fig. \ref{fig:logits-entropy-all}, rows 1-5, last column). This yields many high-confidence false positive (FP) predictions, yielding a marked performance drop at high referral rates. REMEDIS/DAN (Fig. \ref{fig:logits-entropy-all}, last row) mitigates the long tail of the OOD healthy class logit distribution (blue) somewhat, thereby ameliorating referral failures.

\subsection*{E.7. Significance Testing}

We evaluated all models by running the finetuning-evaluation pipeline 6 times, each with different pseudorandom seeds. We perform a two-sided {\it t}-test (uncorrected) to test for statistically significant differences between the best performing models and others. We use {\tt scipy.stats.ttest\_ind}\footnote{\url{https://docs.scipy.org/doc/scipy/reference/generated/scipy.stats.ttest_ind.html}} to find the pairwise p-values between methods, and do not assume equal variance.

\subsection*{E.8. Importance Weighting with lower LR}

Training with IW is known to be challenging for over-parameterized deep networks ~\cite{what2019-iw, xu2021understanding-iw, rethink2020-iw}. The source of the challenge is as follows: in IW, those ID samples which are more similar to the OOD distribution are given more weightage when training the classifier. If the OOD distribution were too different from the ID distribution, it would be easier for the domain predictor to distinguish between the two, leading to very large weights for a few atypical ID samples (that are similar to OOD). This may hurt ID classification performance because typical ID samples could be ignored or given lesser importance during fine tuning, due to their small weights.

We observed that IW methods were especially sensitive to the learning rate, as compared to the other methods. For example, for the Diabetic Retinopathy data, we reported in the main text that the IW methods (ViT+IW and REMEDIS/DAN+IW) suffer OOD referral failures (Fig. \ref{fig:iw-llr}); these results were obtained with the default learning rate ($\alpha = 10^{-3}$). Yet, when we reduced the learning rate (to $\alpha = 10^{-4}$) we find that both IW methods perform superlatively on OOD referral (Fig. \ref{fig:iw-llr}c-d, dashed curves). Although ViT+IW suffers non-monotonicity in the AUROC-referral curve (Fig. \ref{fig:iw-llr}c, dashed orange), both models perform much better than their counterparts with the default learning rates (Fig. \ref{fig:iw-llr}c-d, solid curves). But this improvement comes at a cost: for the lower learning rate both methods exhibit a marked drop in ID referral performance (Fig. \ref{fig:iw-llr}a-b, dashed curves), as compared to their counterparts with the default learning rates (Fig. \ref{fig:iw-llr}a-b, solid curves)

\section*{F. Reproducibility Checklist}

\subsection*{General}
\begin{enumerate}
\item This paper includes a conceptual outline and/or pseudocode description of AI methods introduced \textbf{-- Yes}
\item This paper clearly delineates statements that are opinions, hypothesis, and speculation from objective facts and results \textbf{-- Yes}
\item This paper provides well marked pedagogical references for less-familiar readers to gain background necessary to replicate the paper \textbf{-- Yes}
\end{enumerate}

\subsection*{Theoretical contributions}
\begin{enumerate}
\item All assumptions and restrictions are stated clearly and formally. \textbf{-- N/A}
\item All novel claims are stated formally (e.g., in theorem statements). \textbf{-- N/A}
\item Proofs of all novel claims are included.
\item Proof sketches or intuitions are given for complex and/or novel results. \textbf{-- N/A}
\item Appropriate citations to theoretical tools used are given. \textbf{-- N/A}
\item All theoretical claims are demonstrated empirically to hold. \textbf{-- N/A}
\item All experimental code used to eliminate or disprove claims is included. \textbf{-- N/A}
\end{enumerate}

\subsection*{Datasets}
\begin{enumerate}
\item A motivation is given for why the experiments are conducted on the selected datasets \textbf{-- Yes}

Datasets which we have used are publicly available. Moreover, these datasets are widely studied as benchmarks for domain generalization~\cite{azizi2023} and selective classification~\cite{band2021benchmarking}. 

\item All novel datasets introduced in this paper are included in a data appendix. \textbf{-- N/A}
\item All novel datasets introduced in this paper will be made publicly available upon publication of the paper with a license that allows free usage for research purposes. \textbf{-- N/A}
\item All datasets drawn from the existing literature (potentially including authors’ own previously published work) are accompanied by appropriate citations. \textbf{-- Yes}
\item All datasets drawn from the existing literature (potentially including authors’ own previously published work) are publicly available. \textbf{-- Yes}
\item All datasets that are not publicly available are described in detail, with explanation why publicly available alternatives are not scientifically satisficing. \textbf{-- N/A}
\end{enumerate}

\subsection*{Computational experiments}
\begin{enumerate}
\item Any code required for pre-processing data is included in the appendix. \textbf{-- Yes}
\item All source code required for conducting and analyzing the experiments is included in a code appendix. \textbf{-- Yes}
\item All source code required for conducting and analyzing the experiments will be made publicly available upon publication of the paper with a license that allows free usage for research purposes. \textbf{-- Yes}
\item All source code implementing new methods have comments detailing the implementation, with references to the paper where each step comes from  \textbf{-- N/A}
\item If an algorithm depends on randomness, then the method used for setting seeds is described in a way sufficient to allow replication of results. \textbf{-- Yes}
\item This paper specifies the computing infrastructure used for running experiments (hardware and software), including GPU/CPU models; amount of memory; operating system; names, and versions of relevant software libraries and frameworks. \textbf{-- Yes}
\item This paper formally describes evaluation metrics used and explains the motivation for choosing these metrics. \textbf{-- Yes}
\item This paper states the number of algorithm runs used to compute each reported result. \textbf{-- Yes}
\item Analysis of experiments goes beyond single-dimensional summaries of performance (e.g., average; median) to include measures of variation, confidence, or other distributional information. \textbf{-- Yes}
\item The significance of any improvement or decrease in performance is judged using appropriate statistical tests (e.g., Wilcoxon signed-rank). \textbf{-- Yes}
\item This paper lists all final (hyper-)parameters used for each model/algorithm in the paper’s experiments. \textbf{-- Yes}
\item This paper states the number and range of values tried per (hyper-) parameter during development of the paper, along with the criterion used for selecting the final parameter setting. \textbf{-- Yes}
\end{enumerate}

\end{document}